\documentclass[sigconf]{acmart}
\usepackage{bbold}

\AtBeginDocument{%
  \providecommand\BibTeX{{%
    \normalfont B\kern-0.5em{\scshape i\kern-0.25em b}\kern-0.8em\TeX}}}

\setcopyright{acmcopyright}
\copyrightyear{2022}
\acmYear{2022}

\acmConference[Online '22]{Online '22: ACM Web Conference 2022}{April 25, 2022}{Online}
\acmBooktitle{Online '22: ACM Web Conference 2022,
  April 25, 2022, Online}
\acmPrice{}
\acmISBN{}

\begin{document}

\title{Sparsification and Filtering for Spatial-temporal GNN in Multivariate Time-series}

\author{Yuanrong Wang}
\affiliation{
  \institution{University College London}
  \streetaddress{66-72 Gower Street}
  \city{London}
  \country{UK}}
\email{yuanrong.wang@cs.ucl.ac.uk}

\author{Tomaso Aste}
\authornote{Corresponding author.}
\affiliation{
  \institution{University College London}
  \streetaddress{66-72 Gower Street}
  \city{London}
  \country{UK}}
\email{t.aste@ucl.ac.uk}

\renewcommand{\shortauthors}{Y.Wang and T.Aste}

\begin{abstract}
We propose an end-to-end architecture for multivariate time-series prediction that integrates a spatial-temporal graph neural network with a matrix filtering module. This module generates filtered (inverse) correlation graphs from multivariate time series before inputting them into a GNN. In contrast with existing sparsification methods adopted in graph neural network, our model explicitly leverage time-series filtering to overcome the low signal-to-noise ratio typical of complex systems data. We present a set of experiments, where we predict future sales from a synthetic time-series sales dataset. The proposed spatial-temporal graph neural network displays superior performances with respect to baseline approaches, with no graphical information, and with fully connected, disconnected graphs and unfiltered graphs.

\end{abstract}

\ccsdesc[300]{Computing methodologies~Machine learning approaches}

\keywords{Graph Neural Network, LSTM, GCN, GAT, Attention, Sparse Graph, Correlation Matrix, Time-series Forecasting, Information Filtering Network}


\maketitle

\section{Introduction} \label{intro}

Multivariate time-series prediction is an important, general, challenge in data science and machine learning. 
Several deep learning approaches have been proposed in the literature to address multivariate time-series forecasting. However, while they often perform well at extracting temporal patterns, most of the proposed approaches are not designed to account for the interdependency between time-series. Graph neural network, on the other hand, can model time-series as nodes in a graph to account for dependency. Recent development in spatial-temporal graph neural network has been shown to enable multivariate time-series learning and inference \cite{Khodayar2019SpatioTemporalGD,Kapoor2020ExaminingCF,Wan2019MultivariateTC}.

Applications of multivariate time-series prediction ranges from day-to-day business e.g., sales forecasting, traffic prediction to convoluted topics like bio-statistics and action recognition. It is also one of the cornerstones of modern quantitative finance. At the finest granular level, in finance, modeling the levels of limit order book is a multivariate problem for high-frequency trading with the aim of mid-price prediction \cite{Briola2020DeepLM, Briola2021DeepRL}. For longer-term investment like portfolio management \cite{Markowitz,Wang2021DynamicPO}, prices of assets in a portfolio are usually multivariate time-series. Multivariate time series forecasting methods assume inter-dependencies among dynamically changing variables, which captures systematic trends. Namely, the prediction for each variable not only depends on its historical temporal information, but also the others. Understanding this inter-dependency helps to reveal the underlying dynamics of a larger picture, e.g., the financial market, urban transportation system or the urban distribution of shopping centers. However, its complexity is a key challenge that has been studied for over six decades. 

Intra-series temporal patterns and inter-series correlations are jointly the two cores in multivariate time-series forecasting. Recent advancement in deep learning has enabled strong temporal pattern mining. Recurrent Neural Network (RNN) \cite{Rumelhart1986LearningIR}, Long Short-Term Memory (LSTM) network \cite{Sak2014LongSM}, Gated Recurrent Units (GRU) \cite{Chung2014EmpiricalEO}, Gated Linear Units (GLU) \cite{Dauphin2017LanguageMW} and Temporal Convolution Networks (TCN) \cite{Bai2018AnEE} demonstrate promising results in temporal modelling. However, existing methods fail to exploit latent inter-dependencies and correlation among time-series. Historical attempts have been made to input covariance/correlation structure into neural network. Matrix-based neural networks have been discussed \cite{Gao2017MatrixNN,Cai2006LearningWT,Daniusis2008NeuralNW}, but this approach is not specifically designed for the covariance/correlation matrix, and therefore fails to directly and explicitly address the dependency in the covariance/correlation structure inside the calculation.

Graph is a mathematical structure to model the pairwise relation between objects. The permutation-invariant, local connectivity and compositionality of graphs present a perfect data structure to simulate the correlation/covariance matrix. In fact, network science literature has long been including (sparse) covariance/correlation as a special network for analysis \cite{Chen2018EstimatingLC,Turiel2020SimplicialPO,Procacci2021ForecastingMS,Kojaku2019ConstructingNB, Shen2010CovarianceCM}, and many network properties of covariance/correlation matrix contribute greatly to analytical and predictive tasks in the financial market \cite{Millington2017RobustPR,Yuan2020ImprovedLD,Lee2020OptimalPU}. Recently, graph neural network (GNN) has been leveraged to incorporate the topology structure between entities. Hence, modeling inter-series correlation via graph learning is a natural extension to analyzing covariance/correlation matrix from a network perspective. Each variable from a multivariate time-series is a node in the graph, and the edge represents their latent inter-dependency. By propagating information between neighboring nodes, the graph neural network enables each time-series to be aware of correlated context.

Spatial-temporal graph neural network is the most used network structure for multivariate time-series problems in the literature \cite{Zhao2020TGCNAT,Wu2020ConnectingTD,Cao2020SpectralTG,Sesti2021IntegratingLA}, as the temporal part extracts patterns in each uni-variate series with a LSTM/RNN/GRU, while the spatial part (GNN) models the relationship between series with a pre-defined topology or a graph representation learning algorithm. On one hand, existing GNNs heavily utilizes a pre-defined topology structure which is not explicit in multivariate time-series, and does not reflect the temporal dynamics nature of time-series. On the other hand, many graph representation learning methods focus more on generating node embeddings rather than topological structure, and most of the embeddings depend on a pre-defined topological prior or attention mechanism \cite{Ying2018HierarchicalGR,Zhao2020TGCNAT,Franceschi2019UnsupervisedSR}.

In this paper, we propose an end-to-end framework termed Filtered Sparse Spatial-temporal GNN (FSST-GNN) for sales prediction of 50 products in 10 stores. By integrating modern spatial-temporal GNN with traditional matrix filtering/sparsification methods, we demonstrate the direct use of the (inverse) correlation matrix in GNN. Correlation filtering techniques generate a sparse inverse correlation matrix from multivariate time-series, which can be inverted to a filtered correlation matrix. Both the (inverse) correlation can be used as a pre-defined topological structure or prior for further representation learning. With the designed architecture, we further illustrate that filtered graphs generates a positive impact in multivariate time-series learning, and sparse graphs acts as a contributing prior to guide attention mechanism in GNN.

\section{Related Works}

\subsection{Multivariate Time-series Forecasting}
Time-series forecasting is one of the long-standing key problems in statistics, data science and machine learning. Techniques ranges from traditional pattern recognition to modern machine learning. Univariate time-series forecasting focuses on analyzing independent time-series by extracting temporal patterns based on historical behaviours, e.g., the moving average (MA), the auto-regressive (AR), the auto-regressive moving average (ARMA) and the autoregressive integrated moving average (ARIMA) \cite{Young1972TimeSA}. Modern machine learning model like LSTM has been shown as a good fit to tackle this problem by many literature, e.g., FC-LSTM \cite{Shi2015ConvolutionalLN} and SMF \cite{Zhang2017StockPP}. Multivariate forecasting considers a correlated collection of time-series. The vector auto-regressive model (VAR) and the vector auto-regressive moving average model (VARMA) \cite{Kascha2012ACO,Anderson1978MaximumLE} extend the aforementioned linear models into a multivariate space by capturing the interdependecy between time-series. Early attempts combines convolution neural network (CNN) and recurrent neural network (RNN) to learn local spatial dependencies and temporal patterns \cite{Lai2018ModelingLA,Shih2019TemporalPA}. Further works include state space model in Deep-State \cite{Rangapuram2018DeepSS} and matrix factorization approach in DeepGLO \cite{Sen2019ThinkGA}.

\subsection{Spatial-temporal Graph Neural Network}
Spatial-temporal graph neural network has been proposed recently for multivariate time-series problems. To capture the correlation be between time-series in the spatial component, each time-series is modelled as a node in a graph whereas the edge between every two nodes represents their correlation. Early work applies spatial-temporal GNN for traffic forecasting \cite{Li2018DiffusionCR,Yu2018SpatioTemporalGC,Chen2020MultiRangeAB,Wu2019GraphWF,Zheng2020GMANAG}. Further studies have been extended to other fields, e.g., action recognition \cite{Shi2019TwoStreamAG,Yan2018SpatialTG} and bio-statistics with many interesting works for COVID-19 \cite{LaGatta2021AnEN,Fritz2021CombiningGN,Kapoor2020ExaminingCF}. For financial applications, Matsunaga et al. \cite{Matsunaga2019ExploringGN} is one of the first studies exploring the idea of incorporating company knowledge graphs directly into the predictive model by GNN. Later, Hou et al. \cite{Hou2021STTraderAS} proposed to use a variational autoencoder (VAE) to process stock fundamental information and cluster it into graph structure. This learned adjacency matrix is then fed into a GCN-LSTM for further forecasting. Similar work has been done by Pillay \& Moodley \cite{Pillay2022ExploringGN} with a different model architecture called Graph WaveNet. The most recent advancement is a spatial-temporal GNN for portfolio/asset management proposed by Amudi \cite{Amudi}. They combine a stock sector graph, a correlation graph and a supply-chain graph into one super graph and use the multi-head attention in GAT as a sparsification method to select the meaningful subgraph for prediction. In line with this work, we focus on filtered/sparsified (inverse) correlation graph generated from matrix filtering/sparsification techniques.

\subsection{Correlation Matrix Filtering}
Many computational methods employ sparse approximation techniques to estimate the inverse covariance matrix. 
The sparsification is effective because the least significant components in a covariance matrix are often largely prone to small changes and can lead to instability. 
Sparsified models filters out these insignificant components, and thus improve the model resilience to noise. As correlation is a scaled form of covariance, filtering and sparsification methods are equivalently applicable in both cases.
\subsubsection{Covariance Shrinkage}
A shrinkage algorithm minimizes the ratio between the smallest and the largest eigenvalues of the empirical covariance matrix, which is done by simply shifting every eigenvalue according to a given offset. This approach is equivalent of finding the $L_2$-penalized maximum likelihood estimator of the covariance matrix \cite{Ledoit2003HoneyIS}, which is expressed as a simple convex transformation:
\begin{equation}\label{eq:covariance srhinkage}
\begin{aligned}
    \Sigma_{\text{shrunk}}=(1-\alpha)\hat{\Sigma}+\alpha\frac{\text{Tr}\hat\Sigma}{n}\mathbb{1}
\end{aligned}
\end{equation} 
where $\Sigma_{\text{shrunk}}$ is the shrunk covariance, $\hat{\Sigma}$ is the empirical covariance, $n$ is the number of features in the covariance, $\mathbb{1}$ is the identity matrix and $\alpha$ is the shrinkage coefficient. To optimise the selection of the shrinkage coefficient, Ledoit and Wolf in 2004 \cite{Ledoit2004AWE} proposed to compute $\alpha$ that minimizes the mean square error between the estimated and empirical covariance. With further assumption on the normality of data, Chen et al. in 2010 \cite{Chen2010ShrinkageAF} proposed a better $\alpha$ computation based on minimum mean square error. Further research focuses on large dimension shrinkage \cite{Ledoit2012NonlinearSE,Donoho2018OptimalSO,Couillet2014LargeDA}.
\subsubsection{Graphical Models}
A widely used approach for inverse covariance estimation is based on graph models. Meinshausen and Buhlmann in 2006 \cite{Meinshausen2006HighdimensionalGA} regards the zero entries in the inverse covariance matrix of a multi-variable normal distribution as conditional independence between variables. These structural zeros can thus be obtained through neighborhood selection with LASSO regression by fitting a LASSO to each variable and using the others as predictors. Similar methods that maximizes $L_1$ penalized log-likelihood have been studies by Yuan and Lin \cite{Yuan2007ModelSA} and Banerjee et al. \cite{Banerjee2007ModelST}. In 2008, Friedman et al. \cite{CompNet9} developed an efficient Graphical LASSO that uses $L_1$ norm regularization to control the sparsity in the precision matrix. The sparse inverse covariance matrix can be obtained through minimizing the regularized negative log-likelihood \cite{Mazumder2012TheGL}:
\begin{equation}\label{eq:glasso}
\begin{aligned}
    \Sigma_{\text{glasso}}^{-1}=\min_{{\mathbf \Sigma^{-1}}}( -\log\det{\Sigma^{-1}} + \text{Tr}(\hat\Sigma^{-1}\Sigma^{-1})+\lambda||\Sigma^{-1}||_{1})
\end{aligned}
\end{equation} 
where $\hat\Sigma^{-1}$ is the empirical inverse covariance, $||\Sigma^{-1}||_{1}$ denotes the sum of the absolute values of $\Sigma^{-1}$, and $\lambda$ is the regularization constant, optimised by cross-validation. 
\subsubsection{Information Filtering Network} \label{IFN}
An alternative approach that uses information filtering networks has been shown to deliver better results with lower computational burden and larger interpretability \cite{CompNet7}. In the past few years, information filtering network analysis of complex system data has advanced significantly. It models interactions in a complex system as a network structure of elements (vertices) and interactions (edges). The best-known approach, the Minimum Spanning Tree (MST) was firstly introduced by Boruvka in 1926 \cite{nevsetvril2001otakar} and it can be solved exactly (see \cite{CompNet10} and  \cite{CompNet11} for two common approaches). 
The MST reduces the structure to a connected tree which retains the larger correlations.
To better extract useful information, Tumminello et al. \cite{CompNet3} and Aste and Di Matteo \cite{CompNet4} introduced the use of planar graphs in the Planar Maximally Filtered Graph (PMFG) algorithm. Recent studies have extended the approach to chordal graphs of flexible sparsity \cite{CompNet5, CompNet6}. Research fields ranging from finance \cite{CompNet7} to neural systems \cite{CompNet8} have applied this approach as a powerful tool to understand high dimensional dependency and construct a sparse representation. 
It was shown that, for chordal information filtering networks, such as the Triangulated Maximally Filtered Graph (TMFG) \cite{CompNet5}, one can obtain a sparse precision matrix that is positively definite and has the structure of the network paving the way for a proper $L_0$-norm topological regularization \cite{aste2020topological}. Further study in Maximally Filtered Clique Forest (MFCF) \cite{Massara2019LearningCF} extends the generality of the method by applying it to different sizes of cliques. This approach has proven to be computationally more efficient and stable than Graphical LASSO \cite{CompNet9} and covariance shrinkage methods \cite{Ledoit2003HoneyIS,Ledoit2004AWE,Chen2010ShrinkageAF}, especially when few data points are available \cite{CompNet7, CompNet4}. 

\subsection{Sparse GNN}
Many literature has discussed graph sparsification in GNN. Some, by including regularization, reduce unnecessary edges, which can largely improve the efficiency and efficacy of large-scale graph problems \cite{Calandriello2018ImprovedLG,Chakeri2016SpectralSI}. Some leverage stochastic edge pruning in graphs as a dropout-equivalent regularization to enhance the training process \cite{Rong2020DropEdgeTD,Hasanzadeh2020BayesianGN}. Others train the GNN to learn sparsification as an integrated part before applying it to downstream tasks. NeuralSparse learns to sample k-neighbor subgraph as input for GNN \cite{Zheng2020GMANAG}. Luo proposes to prune task-irrelevant edges \cite{Luo2021LearningTD}. Kim uses the disconnected edges of sparse graphs to guide attention in GAT \cite{Kim2021HowTF}.

\section{Model Implementation}

\begin{figure*}[t]
    \centering
    \includegraphics[width=\textwidth]{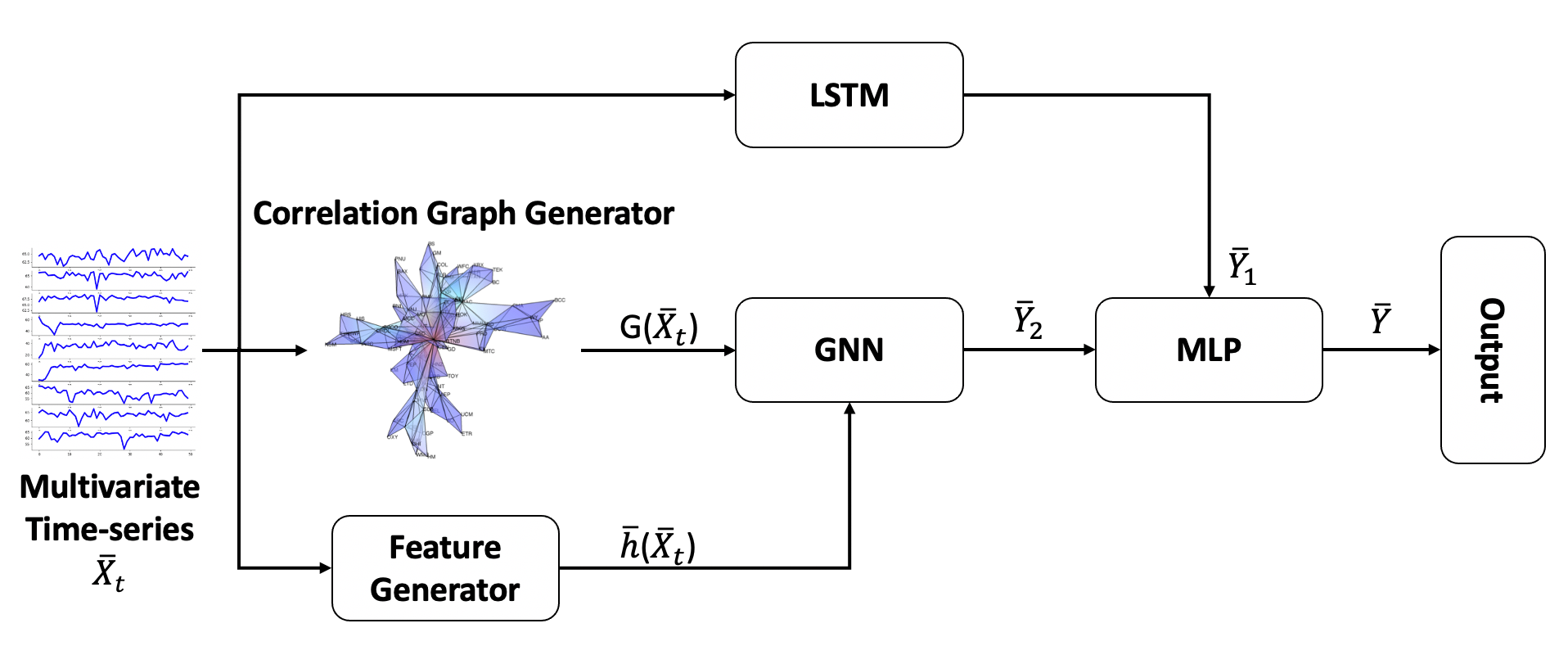}
    \vspace{0.5pt}
    \caption[Architecture]{The overall model architecture.}
         \label{fig:Architecture}
\end{figure*}

We first elaborate on the general framework of our model. As illustrated in Figure \ref{fig:Architecture}, the model consists of 5 main building blocks. A correlation graph generator is able to transform the multivariate time-series into a correlation graph where each node represents a single time-series and each edge between two nodes denotes their correlation. A standard transformation generates a full (inverse) correlation graph with (inverse) correlation edges between each node. In addition, correlation-filtering based transformation generates a full correlation graph with filtered correlation edges, or a sparse inverse correlation graph. We employ covariance shrinkage, graphical models and information filtering network as the three main correlation-filtering based graph generators. The feature generator generates initial input features for each node based on the multivariate time-series. The generated graph and the features from the two generators are then fed into a GNN to learn meaningful node embeddings as the spatial information. Similarly, the multivariate time-series is also fed into a LSTM to extract temporal information. Then, the spatial and temporal information are input in a multi-layer perceptron (MLP) as the read-out layer for the final output, the predicted sales number.

\subsection{Correlation Graph Generator}
\subsubsection{Covariance Shrinkage}
Covariance shrinkage method as described in equation \ref{eq:covariance srhinkage} is equivalently applicable to the correlation matrix. Shrinkage coefficient $\alpha$ is optimized by cross-validation. There is a implementation, {\fontfamily{cmr}\selectfont \textbf{sklearn.covariance.ShrunkCovariance}} Pyhon library \cite{scikit-learn}, which is applied in this experiment. Filtered correlation is then directly transform into a graph. Inverse correlation can be obtained by direct matrix inversion, which is implemented by the {\fontfamily{cmr}\selectfont \textbf{numpy.linalg.inv}} library \cite{2020NumPy-Array}.

\subsubsection{Graphical LASSO}
Graphical LASSO is a graphical model for inverse covariance sparsification, which is epressed in equation \ref{eq:glasso}. We leverage Python's {\fontfamily{cmr}\selectfont \textbf{sklearn.covariance.GraphicalLasso}} library for implementation, and {\fontfamily{cmr}\selectfont \textbf{sklearn.covariance.GraphicalLassoCV}} \cite{scikit-learn} for cross valiation and regularization constant $\lambda$ selection. Graphical LASSO sparsifies an inverse correlation matrix which can be directly transformed into a sparse inverse correlation graph, while a full but filtered correlation graph can be obtained through the matrix inversion of the inverse correlation.

\subsubsection{Maximally Filtered Clique Forest}
We implement Maximally Filtered Clique Forest (MFCF), an information filtering network, for sparse precision matrix filtering. By setting the minimum and maximum clique size to 4, we simplify our solution to a TMFG-equivalent model discussed in Section \ref{IFN}. It generates sparse inverse correlation, which will undergoes similar transformation as Graphical LASSO to obtain inverse correlation and correlation graphs.

\subsection{GNN}
\subsubsection{GCN}
Graph convolution network is proposed by Kipf and Welling in 2017 \cite{Kipf2017SemiSupervisedCW}, which generates embeddings for each node in the graph. It takes original features in each node as the initial embeddings, then aggregates neighboring feature representations and updates the node embeddings through a message-passing like network with the adjacency matrix, which can be expressed as:
\begin{equation}\label{GCN_adjacency}
\begin{aligned}
    h_i=\sum^{N}_{j}{(\frac{\phi(W^\top A_{ij})}{d_j})\times h_j}
\end{aligned}
\end{equation}
where $h_i$ is the node embedding, $h_j$ is the neighboring node embedding, $W$ is a learnable parameter, $\phi$ is non-linear activation, $A_{ij}$ is the adjacency matrix and $d_j$ is the degree of node $j$ for normalization. 

In the experiments, we have replaced the graph information, adjacency matrix, expressed in equation \ref{GCN_adjacency} by (inverse) correlation matrix, adjacency matrix and Laplacian matrix obtained by thresholding the (inverse) correlation matrix. The empirical results suggest the superiority by simply employing the (inverse) correlation matrix. It can be seen as weighted adjacency matrix, where correlation coefficients are naturally scaled/normalized. Therefore, the graph convolution can be re-expressed as:
\begin{equation}\label{GCN_corr}
\begin{aligned}
    h_i=\sum^{N}_{j}{\phi(W^\top C_{ij})\times h_j}
\end{aligned}
\end{equation}
where $C_{ij}$ is the (inverse) correlation matrix, and all the other parameters are previously defined.

\subsubsection{GAT}
The implementation of GCN limits the model to be used only with static graphs. The embedding update is static across time, which assumes non-stationarity in time-series. Graph attention network uses masked multi-head attention mechanism to solve this issue by dynamically assigning attention coefficients between nodes. The normalized attention coefficient $a_{ij}$ is computed for nodes $i$ and $j$ based on their features (embeddings):
\begin{equation}\label{eq:GAT attention}
\begin{aligned}
    e_{i,j}= & a(Wh_i,Wh_j)\\
    = & \textrm{LeakyReLU} (\hat{a}^\top[Wh_i||Wh_j]) \\
\end{aligned}
\end{equation}
where $W$ is the learnable linear transformation weight matrix to transform node features, $h$, into lower dimensional representations, $a$ is the attention mechanism to perform self-attention on each node and $||$ represents concatenation operation. We normalize the attention coefficient $e_{ij}$ by a softmax function:
\begin{equation}\label{eq:GAT normalized attention}
    \begin{aligned}
        \alpha_{i,j}=&\textrm{softmax}_j (e_{i,j})\\
        =&\frac{\exp{e_{i,j}}}{\sum_{k\in\mathcal{N}_i}\exp{e_{i,k}}} \\
        =& \frac{\exp{(\textrm{LeakyReLU}(\hat{a}^\top[Wh_i||Wh_j]))}}{\sum_{k\in\mathcal{N}_i} \exp{(\textrm{LeakyReLU}(\hat{a}^\top[Wh_i||Wh_k]))}}
    \end{aligned}
\end{equation}

The multi-head attention mechanism has been proposed by Vaswani et al. \cite{Vaswani2017AttentionIA} which demonstrates superior and robust performance in network training. GAT incorporates the masked multi-head attention where attention is only computed between neighboring nodes, and the output feature representation is expressed as:
\begin{equation}\label{eq:GAT multi-head aggregation}
     h_i = \phi(\frac{1}{k}\sum_{k=1}^K(\sum_{j\in\mathcal{N}_i} \alpha_{i,j}W^kh_j))
\end{equation}
where in each level of attention, the representation embedding is updated by a learnable parameter $W$ and the attention coefficient matrix $a_{i,j}$, then the final representation is averaged between the $K$ multi-head attention layers and applied a non-linearity $\phi$.

\begin{table*}[h!]
    \centering
    \begin{tabular}{|c|c|c c c|}
    \toprule
    Graph      &     Filtering &  RMSE & MAE & MAPE    \\
    \midrule
    \multicolumn{5}{|c|}{FSST-GNN (GCN)}  \\
    \midrule
    Cor &    Empirical &  
    10.12 $\pm$ 0.53 & 7.77 $\pm$ 0.39 & 17.28$\%\pm$1.18$\%$ \\

    Cor &    Shrinkage &  
     9.99 $\pm$ 0.17 & 7.72 $\pm$ 0.09 & 17.34$\%\pm$0.61$\%$ \\

    Cor &    GLASSO &  
     \textbf{9.76 $\pm$ 0.47} & \textbf{7.52 $\pm$ 0.39} & \textbf{16.96$\%\pm$1.49$\%$} \\

    Cor &    MFCF &  
     9.80 $\pm$ 0.61 & 7.66 $\pm$ 0.43 & 17.27$\%\pm$1.13$\%$  \\
    \midrule
    
    Inv Cor &  Empirical  &  
    11.74 $\pm$ 0.99 & 8.69 $\pm$ 0.46 & 20.08$\%\pm$0.90$\%$  \\

    Inv Cor &    Shrinkage &  
    10.59 $\pm$ 1.04 & 8.26 $\pm$ 0.78 & 19.15$\%\pm$2.21$\%$  \\

    Inv Cor &    GLASSO &  
    \underline{\textbf{9.67 $\pm$ 0.41}} & \underline{\textbf{7.55 $\pm$ 0.34}} & \underline{\textbf{17.51$\%\pm$1.54$\%$}} \\

    Inv Cor &    MFCF &  
    10.07 $\pm$ 0.59 & 7.99 $\pm$ 0.44 & 17.87$\%\pm$1.11$\%$ \\
    \midrule
    
    Zeros &    / &  
    13.51 $\pm$ 0.16 & 10.28 $\pm$ 0.12 & 22.04$\%\pm$1.01$\%$  \\

    Ones &    / &  
    12.33 $\pm$ 1.12 & 10.00 $\pm$ 0.98 & 24.76$\%\pm$2.33$\%$  \\

    Identity  &    / &  
    11.78 $\pm$ 0.52 & 9.23 $\pm$ 0.46 & 21.01$\%\pm$1.78$\%$  \\
    \midrule
    \multicolumn{5}{|c|}{LSTM}  \\
    \midrule
    / &    / &  16.34 $\pm$ 0.44 & 12.56 $\pm$ 0.25 & 26.40$\%\pm$0.98$\%$ \\
    \bottomrule

    \end{tabular}
    \vspace{0.5pt}
    \caption{Summary of forecasting results with different models, graphs and filtering methods. Highlighted in bold are the optimal RMSE, MAE and MAPE in each graph, and underlined is the absolute optimal results in the table. A LSTM results is attached as the baseline. }
    \label{Tab:comparison table}
\end{table*}

\section{Experiments}
\subsection{Setup}
We test our model on a Kaggle playground code competition, Store Item Demand Forecasting Challenge \cite{kaggle}. The dataset consists of 5-year sales time-series data of 50 products in 10 different stores. For simplicity, we re-formulate the problem as 50 mini-problems, each focuses on 1 product in 10 different stores. At each time-stamp, the temporal component regresses each of the 10 time-series individually based on its historical value. The dependency between them is reflected by the final embeddings generated from the spatial component. The outputs from each component are subsequently concatenated and, by a read-out layer, to generate final daily forecasting for the product. We assume stationarity in the time-series, therefore, we separate the training and testing data as the 80\% and 20\% of the raw dataset.

The temporal component of the FSST-GNN is a LSTM, which has an input size of ($t_{lb}$, 10) where $t_{lb}=14$ is the look back window size of historical sales, and 10 is the number of different stores. The feature generator produces node features (initial embeddings). We employ the four moments (mean, standard deviation, skewness and kurtosis) of the sales time-series distribution based on each 14-day look back window. The correlation graph generator generates a graph with edge represents the correlation between any two of the 14-day sample time-series in the 10 stores. Then, the generated node features and edges are input into the GNN. In the experiments, a GCN and a GAT have been used as the spatial component.

To understand the effect of filtering and sparsification for multivariate time-series graph learning, we perform 4 sets of experiment: 1) FSST-GNN (GCN) on different filtered correlation graphs; 2) FSST-GNN (GCN) on different filtered inverse correlation graphs; 3) FSST-GNN (GCN) on GLASSO-filtered and MFCF-filtered inverse correlation graph with different levels of sparsity; and 4) FSST-GNN (GAT) on GLASSO-filtered and MFCF-filtered inverse correlation graph with different levels of sparsity. Each experiment has been re-computed 10 times with different random seeds, and the final results is averaged for statistical robustness.

\begin{table*}[h!]
    \centering
    \begin{tabular}{|c|c c c |c | ccc|}
    \toprule
   Sparsity &  RMSE & MAE & MAPE  & Sparsity &  RMSE & MAE & MAPE\\
   \midrule
   \multicolumn{4}{|c|}{GLASSO} & \multicolumn{4}{c|}{MFCF} \\
    \midrule
     77.2\% &  10.24 $\pm$ 0.61 & 8.03$\pm$ 0.47 & 18.89$\%\pm$1.35$\%$ & 
     76.6\% & 11.35 $\pm$ 0.76 & 8.76 $\pm$ 0.61 & 20.36$\%\pm$1.97$\%$\\
    \midrule
     71.0\% &  10.34 $\pm$ 0.79 & 8.05 $\pm$ 0.58 & 18.66$\%\pm$1.29$\%$ & 
     72.3\% & 10.23 $\pm$ 0.26 & 8.06 $\pm$ 0.44 & 18.13$\%\pm$1.14$\%$\\
    \midrule
     66.6\% &  9.80 $\pm$ 0.41 & 7.66 $\pm$ 0.33 & 17.75$\%\pm$0.87$\%$ & 
     68.6\% & 10.68 $\pm$ 0.80 & 8.32 $\pm$ 0.52 & 19.56$\%\pm$1.30$\%$\\
    \midrule
     60.0\% &  \underline{\textbf{9.67 $\pm$ 0.41}} & \underline{\textbf{7.55 $\pm$ 0.34}} & \underline{\textbf{17.51$\%\pm$1.54$\%$}}  & 
     61.3\% & \textbf{10.07 $\pm$ 0.59} & \textbf{7.99 $\pm$ 0.44} & \textbf{17.87$\%\pm$1.11$\%$}\\
    \midrule
     56.5\% &  9.86 $\pm$ 0.58 & 7.74 $\pm$ 0.53 & 18.24$\%\pm$1.66$\%$ & 
     58.0\% & 10.19 $\pm$ 0.41 & 8.12 $\pm$ 0.29 & 18.29$\%\pm$0.49$\%$\\
    \midrule
     51.3\% &  10.06 $\pm$ 0.57 & 7.86 $\pm$ 0.45 & 17.68$\%\pm$1.20$\%$ & 
     54.8\% & 10.31 $\pm$ 0.52 & 8.09 $\pm$ 0.39 & 18.22$\%\pm$1.26$\%$\\
    \midrule
     43.3\% &  9.99 $\pm$ 0.27 & 7.88 $\pm$ 0.19 & 17.60$\%\pm$0.47$\%$ & 
     43.7\% & 10.75 $\pm$ 0.68 & 8.16 $\pm$ 0.40 & 18.44$\%\pm$0.55$\%$\\
    
    \bottomrule

    \end{tabular}
    \vspace{0.5pt}
    \caption{Summary of forecasting results of FSST-GNN (GCN) with different filtering methods and sparsity on inverse correlation graph. Highlighted in bold are the optimal RMSE, MAE and MAPE in each model-sparsity combination, and underlined is the absolute optimal results in the table.}
    \label{Tab:GCN_sparsity}
\end{table*}

\begin{table*}[h!]
    \centering
    \begin{tabular}{|c|c c c |c | ccc|}
    \toprule
   Sparsity &  RMSE & MAE & MAPE  & Sparsity &  RMSE & MAE & MAPE\\
   \midrule
   \multicolumn{4}{|c|}{GLASSO} & \multicolumn{4}{c|}{MFCF} \\
    \midrule
     77.2\% &  9.88 $\pm$ 0.59 & 7.58 $\pm$ 0.43 & 16.17$\%\pm$0.53$\%$ & 
     76.6\% & 10.47 $\pm$ 0.75 & 8.09 $\pm$ 0.68 & 17.61$\%\pm$2.64$\%$\\
    \midrule
     71.0\% &  10.03 $\pm$ 0.65 & 7.73 $\pm$ 0.52 & 16.56$\%\pm$1.07$\%$ & 
     72.3\% & 10.27 $\pm$ 0.62 & 7.86 $\pm$ 0.46 & 17.24$\%\pm$0.99$\%$\\
    \midrule
     66.6\% &  9.63 $\pm$ 0.36 & 7.42 $\pm$ 0.25 & 15.82$\%\pm$0.25$\%$ & 
     68.6\% & 9.90 $\pm$ 1.17 & 7.60 $\pm$ 0.95 & 16.01$\%\pm$1.78$\%$\\
    \midrule
     60.0\% & \textbf{ 9.58 $\pm$ 0.31} & \textbf{7.37 $\pm$ 0.20} & \textbf{15.62$\%\pm$0.26$\%$}  & 
     61.3\% & \underline{\textbf{9.46 $\pm$ 0.68}} & \underline{\textbf{7.31 $\pm$ 0.45}} & \underline{\textbf{15.44$\%\pm$0.19$\%$}}\\
    \midrule
     56.5\% &  9.64 $\pm$ 0.34 & 7.41 $\pm$ 0.23 & 15.80$\%\pm$0.47$\%$ & 
     58.0\% & 9.65 $\pm$ 0.46 & 7.53 $\pm$ 0.32 & 15.63$\%\pm$0.54$\%$\\
    \midrule
     51.3\% &  9.75 $\pm$ 0.33 & 7.55 $\pm$ 0.31 & 17.28$\%\pm$1.18$\%$ & 
     54.8\% & 9.81 $\pm$ 0.53 & 7.53 $\pm$ 0.37 & 16.03$\%\pm$0.54$\%$\\
    \midrule
     43.3\% &  10.12 $\pm$ 0.53 & 7.77 $\pm$ 0.39 & 17.28$\%\pm$1.18$\%$ & 
     43.7\% & 9.90 $\pm$ 0.38 & 7.63 $\pm$ 0.30 & 16.31$\%\pm$0.92$\%$\\
    
    \bottomrule

    \end{tabular}
    \vspace{0.5pt}
    \caption{Summary of forecasting results of FSST-GNN (GAT) with different filtering methods and sparsity on inverse correlation graph. Highlighted in bold are the optimal RMSE, MAE and MAPE in each model-sparsity combination, and underlined is the absolute optimal results in the table.}
    \label{Tab:GAT_sparsity}
\end{table*}
\subsection{Results}

We compute the root mean square error (RMSE), mean average error (MAE) and mean average percentage error (MAPE) of the predicted sales number of all 50 products in 10 stores with the ground truth label in Table \ref{Tab:comparison table} as the evaluation matrix to analyze the effectiveness of filtering methods over FSST-GNN with GCN on the correlation and inverse correlation graph respectively. Since all filtering methods are parametric, the table reports the optimal results from covariance shrinkage (Shrinkage), graphical LASSO (GLASSO) and MFCF, which are obtained through grid-search. We also include a fully connected graph of a matrix of ones, two fully disconnected graphs of a matrix of zeros and an identity matrix as benchmarks for comparison. In addition, a plain LSTM is also presented as the baseline model where no graphical/spatial information is input.

In Table \ref{Tab:comparison table}, it is evident that all FSST-GNN (GCN) outperforms the plain LSTM, which confirms the efficacy of considering the spatial information in multivariate time-series problems. Other benchmarks of fully connected/disconnected graphs are also presented, and their results in all three measurements are effectively inferior to any (inverse) correlation based graph methods. These results further assert the information gain from meaningful spatial graphs.

Highlighted in each column of Table \ref{Tab:comparison table} are the best results in first two experiment: 1) FSST-GNN (GCN) on correlation graph; 2) FSST-GNN (GCN) on inverse correlation graph. In correlation graph cases, a filtered correlation graphs demonstrate superior results than the original Empirical correlation graph. Both MFCF and GLASSO filtering are operated on the inverse correlation for sparsification, and then inverted back to a full correlation graphs, while Shrinkage operates directly on correlation. Therefore, the superior results in MFCF and GLASSO than Shrinkage may suggest a stronger filtering effect behind graph/network-based methods, and inversion does not affect filtering.

To understand the effect in filtering and sparsification, results from the same setup with inverse correlation graphs are compared, where full and Shrinkage-filtered inverse correlation graphs are full graphs and GLASSO-filtered and MFCF-filtered graphs are sparse graphs. In this case, Shrinkage filters a correlation and inverts it to an inverse correlation. Comparably to the correlation graph case, Shrinkage consistently yields better result than the Empirical, which further validates that the filtering mechanism is hardly impacted by inversion operation. Furthermore, we observe even more significant results from two sparse graphs filtered by GLASSO and MFCF. This advantage could possibly come from both the filtering, the sparsification, as well as their combined effect. To further investigate the sole efficacy of sparsification, we perform the third and fourth sets of experiments: 3) FSST-GNN (GCN) on GLASSO-filtered and MFCF-filtered inverse correlation graph with different levels of sparsity; and 4) FSST-GNN (GAT) on GLASSO-filtered and MFCF-filtered inverse correlation graph with different levels of sparsity.

Presented in Table \ref{Tab:GCN_sparsity} and Table \ref{Tab:GAT_sparsity} are the results with different levels of sparsity. We select the parameter to match the sparsity level between GLASSO and MFCF for comparison. It is seen that at around $60\%$ sparsity, the highlighted best results are achieved for both MFCF and GLASSO in FSST-GNN (GCN) and FSST-GNN (GAT) models. Moreover, as the sparsity deviates away from this local minimum, the three errors start to increase, which may suggest an optimal sparsity structure of the inverse correlation graph in our experimental case. In addition, this optimal structure is independent of the chosen model. Furthermore, as illustrated in equation \ref{eq:GAT multi-head aggregation}, GAT by default does not account for edge weights in weighted graphs (correlation graphs) as GCN. Hence, the sparse inverse correlation graph serves as a thresholded adjacency matrix, where 0 entries are interpreted as disconnection between nodes. Then, attention, which is only calculated between linked nodes, acts as the edge weights. Namely, the superior performance in Table \ref{Tab:GCN_sparsity} is a mixture of filtering and sparsity, but the performance in Table \ref{Tab:GAT_sparsity} is merely determined by the sparsity of the input graph without filtering mechanism.

\section{Conclusion}

Literature has presented many GNN-based graph sparsification methods. However, none of them explicitly addresses the filtering and sparsification from a time-series perspective. In small sample time-series problems, especially in finance, graph structure learning models, e.g., graph representation learning, are highly prone to noise. In this paper, we design an end-to-end filtered sparse spatial-temporal graph neural network for time-series forecasting. Our model leverages and integrates traditional matrix filtering methods with modern graph neural networks to achieve robust results, and show the use of a simple and efficient architecture. We employ three different matrix filtering methods, covariance shrinkage, graphical LASSO and information filtering network-maximally filtered clique forest to show a positive gain in graph filtering to graph learning. The results from the three methods surpass all of the benchmark approaches, including a LSTM with no graphical information, the same FSST-GNN architecture with fully connected, disconnected graphs and unfiltered graphs.

In the experiments, we found the sparse graph in GAT serves only as an indication of which pairs of node require attention calculation, and the advantages from sparsity are significant. The filtered correlation matrix in GCN is interpreted and used as a weighted adjacency matrix for direct graph convolution, where the efficacy of filtering is also obvious. Furthermore, the optimal combined effect of filtering and sparsification in FSST-GNN (GCN) with inverse correlation implies the two contributing factors are complementary. Therefore, by incorporating weighted graphs in GAT like Grassia \& Mangioni \cite{Grassia2022wsGATWA}, we may further improve the performance of attention-based graph neural networks.
 
Current work is based on a synthetic dataset from a Kaggle competition for sales prediction. Further work will be applied with real world financial data for practical problems, e.g., portfolio optimization, risk management and price forecasting. The temporal and spatial component of the current architecture are designed to compute in parallel and combined in the end. Therefore, temporal information does not directly contribute to the spatial filtered graph generation or graph node feature generation. In the next phase of this study, we aim to develop a stacked architecture, where temporal signals contribute to spatial graph filtering/sparsification.

\newpage
\



\newpage

\begin{thebibliography}{85}


\ifx \showCODEN    \undefined \def \showCODEN     #1{\unskip}     \fi
\ifx \showDOI      \undefined \def \showDOI       #1{#1}\fi
\ifx \showISBNx    \undefined \def \showISBNx     #1{\unskip}     \fi
\ifx \showISBNxiii \undefined \def \showISBNxiii  #1{\unskip}     \fi
\ifx \showISSN     \undefined \def \showISSN      #1{\unskip}     \fi
\ifx \showLCCN     \undefined \def \showLCCN      #1{\unskip}     \fi
\ifx \shownote     \undefined \def \shownote      #1{#1}          \fi
\ifx \showarticletitle \undefined \def \showarticletitle #1{#1}   \fi
\ifx \showURL      \undefined \def \showURL       {\relax}        \fi
\providecommand\bibfield[2]{#2}
\providecommand\bibinfo[2]{#2}
\providecommand\natexlab[1]{#1}
\providecommand\showeprint[2][]{arXiv:#2}

\bibitem[\protect\citeauthoryear{Anderson}{Anderson}{1978}]%
        {Anderson1978MaximumLE}
\bibfield{author}{\bibinfo{person}{Theodore~W. Anderson}.}
  \bibinfo{year}{1978}\natexlab{}.
\newblock \showarticletitle{Maximum Likelihood Estimation for Vector
  Autoregressive Moving Average Models}.
\newblock


\bibitem[\protect\citeauthoryear{Aste}{Aste}{2020}]%
        {aste2020topological}
\bibfield{author}{\bibinfo{person}{Tomaso Aste}.}
  \bibinfo{year}{2020}\natexlab{}.
\newblock \showarticletitle{Topological regularization with information
  filtering networks}.
\newblock \bibinfo{journal}{\emph{arXiv preprint arXiv:2005.04692}}
  (\bibinfo{year}{2020}).
\newblock


\bibitem[\protect\citeauthoryear{Aste and Matteo}{Aste and Matteo}{2017}]%
        {CompNet4}
\bibfield{author}{\bibinfo{person}{T. Aste} {and} \bibinfo{person}{T. Matteo}.}
  \bibinfo{year}{2017}\natexlab{}.
\newblock \showarticletitle{Sparse Causality Network Retrieval from Short Time
  Series}.
\newblock \bibinfo{journal}{\emph{Complex.}}  \bibinfo{volume}{2017}
  (\bibinfo{year}{2017}), \bibinfo{pages}{4518429:1--4518429:13}.
\newblock


\bibitem[\protect\citeauthoryear{Bai, Kolter, and Koltun}{Bai
  et~al\mbox{.}}{2018}]%
        {Bai2018AnEE}
\bibfield{author}{\bibinfo{person}{Shaojie Bai}, \bibinfo{person}{J.~Zico
  Kolter}, {and} \bibinfo{person}{Vladlen Koltun}.}
  \bibinfo{year}{2018}\natexlab{}.
\newblock \showarticletitle{An Empirical Evaluation of Generic Convolutional
  and Recurrent Networks for Sequence Modeling}.
\newblock \bibinfo{journal}{\emph{ArXiv}}  \bibinfo{volume}{abs/1803.01271}
  (\bibinfo{year}{2018}).
\newblock


\bibitem[\protect\citeauthoryear{Banerjee, Ghaoui, and d'Aspremont}{Banerjee
  et~al\mbox{.}}{2007}]%
        {Banerjee2007ModelST}
\bibfield{author}{\bibinfo{person}{Onureena Banerjee},
  \bibinfo{person}{Laurent~El Ghaoui}, {and} \bibinfo{person}{Alexandre
  d'Aspremont}.} \bibinfo{year}{2007}\natexlab{}.
\newblock \showarticletitle{Model Selection Through Sparse Maximum Likelihood
  Estimation}.
\newblock \bibinfo{journal}{\emph{ArXiv}}  \bibinfo{volume}{abs/0707.0704}
  (\bibinfo{year}{2007}).
\newblock


\bibitem[\protect\citeauthoryear{Barfuss, Massara, Di~Matteo, and Aste}{Barfuss
  et~al\mbox{.}}{2016}]%
        {CompNet7}
\bibfield{author}{\bibinfo{person}{Wolfram Barfuss},
  \bibinfo{person}{Guido~Previde Massara}, \bibinfo{person}{T. Di~Matteo},
  {and} \bibinfo{person}{Tomaso Aste}.} \bibinfo{year}{2016}\natexlab{}.
\newblock \showarticletitle{Parsimonious modeling with information filtering
  networks}.
\newblock \bibinfo{journal}{\emph{Physical Review E}} \bibinfo{volume}{94},
  \bibinfo{number}{6} (\bibinfo{date}{Dec} \bibinfo{year}{2016}).
\newblock
\showISSN{2470-0053}
\urldef\tempurl%
\url{https://doi.org/10.1103/physreve.94.062306}
\showDOI{\tempurl}


\bibitem[\protect\citeauthoryear{Briola, Turiel, and Aste}{Briola
  et~al\mbox{.}}{2020}]%
        {Briola2020DeepLM}
\bibfield{author}{\bibinfo{person}{Antonio Briola}, \bibinfo{person}{Jeremy~D.
  Turiel}, {and} \bibinfo{person}{Tomaso Aste}.}
  \bibinfo{year}{2020}\natexlab{}.
\newblock \showarticletitle{Deep Learning Modeling of the Limit Order Book: A
  Comparative Perspective}.
\newblock \bibinfo{journal}{\emph{ERN: Other Econometrics: Econometric \&
  Statistical Methods - Special Topics (Topic)}} (\bibinfo{year}{2020}).
\newblock


\bibitem[\protect\citeauthoryear{Briola, Turiel, Marcaccioli, and Aste}{Briola
  et~al\mbox{.}}{2021}]%
        {Briola2021DeepRL}
\bibfield{author}{\bibinfo{person}{Antonio Briola}, \bibinfo{person}{Jeremy~D.
  Turiel}, \bibinfo{person}{Riccardo Marcaccioli}, {and}
  \bibinfo{person}{Tomaso Aste}.} \bibinfo{year}{2021}\natexlab{}.
\newblock \showarticletitle{Deep Reinforcement Learning for Active High
  Frequency Trading}.
\newblock \bibinfo{journal}{\emph{ArXiv}}  \bibinfo{volume}{abs/2101.07107}
  (\bibinfo{year}{2021}).
\newblock


\bibitem[\protect\citeauthoryear{Cai, He, and Han}{Cai et~al\mbox{.}}{2006}]%
        {Cai2006LearningWT}
\bibfield{author}{\bibinfo{person}{Deng Cai}, \bibinfo{person}{Xiaofei He},
  {and} \bibinfo{person}{Jiawei Han}.} \bibinfo{year}{2006}\natexlab{}.
\newblock \showarticletitle{Learning with Tensor Representation}.
\newblock


\bibitem[\protect\citeauthoryear{Calandriello, Koutis, Lazaric, and
  Valko}{Calandriello et~al\mbox{.}}{2018}]%
        {Calandriello2018ImprovedLG}
\bibfield{author}{\bibinfo{person}{Daniele Calandriello},
  \bibinfo{person}{Ioannis Koutis}, \bibinfo{person}{Alessandro Lazaric}, {and}
  \bibinfo{person}{Michal Valko}.} \bibinfo{year}{2018}\natexlab{}.
\newblock \showarticletitle{Improved Large-Scale Graph Learning through Ridge
  Spectral Sparsification}. In \bibinfo{booktitle}{\emph{ICML}}.
\newblock


\bibitem[\protect\citeauthoryear{Cao, Wang, Duan, Zhang, Zhu, Huang, Tong, Xu,
  Bai, Tong, and Zhang}{Cao et~al\mbox{.}}{2020}]%
        {Cao2020SpectralTG}
\bibfield{author}{\bibinfo{person}{Defu Cao}, \bibinfo{person}{Yujing Wang},
  \bibinfo{person}{Juanyong Duan}, \bibinfo{person}{Ce Zhang},
  \bibinfo{person}{Xia Zhu}, \bibinfo{person}{Congrui Huang},
  \bibinfo{person}{Yunhai Tong}, \bibinfo{person}{Bixiong Xu},
  \bibinfo{person}{Jing Bai}, \bibinfo{person}{Jie Tong}, {and}
  \bibinfo{person}{Qi Zhang}.} \bibinfo{year}{2020}\natexlab{}.
\newblock \showarticletitle{Spectral Temporal Graph Neural Network for
  Multivariate Time-series Forecasting}.
\newblock \bibinfo{journal}{\emph{ArXiv}}  \bibinfo{volume}{abs/2103.07719}
  (\bibinfo{year}{2020}).
\newblock


\bibitem[\protect\citeauthoryear{Chakeri, Farhidzadeh, and Hall}{Chakeri
  et~al\mbox{.}}{2016}]%
        {Chakeri2016SpectralSI}
\bibfield{author}{\bibinfo{person}{Alireza Chakeri}, \bibinfo{person}{Hamidreza
  Farhidzadeh}, {and} \bibinfo{person}{Lawrence~O. Hall}.}
  \bibinfo{year}{2016}\natexlab{}.
\newblock \showarticletitle{Spectral sparsification in spectral clustering}.
\newblock \bibinfo{journal}{\emph{2016 23rd International Conference on Pattern
  Recognition (ICPR)}} (\bibinfo{year}{2016}), \bibinfo{pages}{2301--2306}.
\newblock


\bibitem[\protect\citeauthoryear{Chen, Kang, Xing, Zhao, and Milton}{Chen
  et~al\mbox{.}}{2018}]%
        {Chen2018EstimatingLC}
\bibfield{author}{\bibinfo{person}{Shuo Chen}, \bibinfo{person}{Jian Kang},
  \bibinfo{person}{Yishi Xing}, \bibinfo{person}{Yunpeng Zhao}, {and}
  \bibinfo{person}{Don Milton}.} \bibinfo{year}{2018}\natexlab{}.
\newblock \showarticletitle{Estimating large covariance matrix with network
  topology for high-dimensional biomedical data}.
\newblock \bibinfo{journal}{\emph{Comput. Stat. Data Anal.}}
  \bibinfo{volume}{127} (\bibinfo{year}{2018}), \bibinfo{pages}{82--95}.
\newblock


\bibitem[\protect\citeauthoryear{Chen, Chen, Xie, Cao, Gao, and Feng}{Chen
  et~al\mbox{.}}{2020}]%
        {Chen2020MultiRangeAB}
\bibfield{author}{\bibinfo{person}{Weiqiu Chen}, \bibinfo{person}{Ling Chen},
  \bibinfo{person}{Yu Xie}, \bibinfo{person}{Wei Cao}, \bibinfo{person}{Yusong
  Gao}, {and} \bibinfo{person}{Xiaojie Feng}.} \bibinfo{year}{2020}\natexlab{}.
\newblock \showarticletitle{Multi-Range Attentive Bicomponent Graph
  Convolutional Network for Traffic Forecasting}.
\newblock \bibinfo{journal}{\emph{ArXiv}}  \bibinfo{volume}{abs/1911.12093}
  (\bibinfo{year}{2020}).
\newblock


\bibitem[\protect\citeauthoryear{Chen, Wiesel, Eldar, and Hero}{Chen
  et~al\mbox{.}}{2010}]%
        {Chen2010ShrinkageAF}
\bibfield{author}{\bibinfo{person}{Yilun Chen}, \bibinfo{person}{Ami Wiesel},
  \bibinfo{person}{Yonina~C. Eldar}, {and} \bibinfo{person}{Alfred~O. Hero}.}
  \bibinfo{year}{2010}\natexlab{}.
\newblock \showarticletitle{Shrinkage Algorithms for MMSE Covariance
  Estimation}.
\newblock \bibinfo{journal}{\emph{IEEE Transactions on Signal Processing}}
  \bibinfo{volume}{58} (\bibinfo{year}{2010}), \bibinfo{pages}{5016--5029}.
\newblock


\bibitem[\protect\citeauthoryear{Chung, Çaglar G{\"u}lçehre, Cho, and
  Bengio}{Chung et~al\mbox{.}}{2014}]%
        {Chung2014EmpiricalEO}
\bibfield{author}{\bibinfo{person}{Junyoung Chung}, \bibinfo{person}{Çaglar
  G{\"u}lçehre}, \bibinfo{person}{Kyunghyun Cho}, {and}
  \bibinfo{person}{Yoshua Bengio}.} \bibinfo{year}{2014}\natexlab{}.
\newblock \showarticletitle{Empirical Evaluation of Gated Recurrent Neural
  Networks on Sequence Modeling}.
\newblock \bibinfo{journal}{\emph{ArXiv}}  \bibinfo{volume}{abs/1412.3555}
  (\bibinfo{year}{2014}).
\newblock


\bibitem[\protect\citeauthoryear{Couillet and Mckay}{Couillet and
  Mckay}{2014}]%
        {Couillet2014LargeDA}
\bibfield{author}{\bibinfo{person}{Romain Couillet} {and}
  \bibinfo{person}{Matthew~R. Mckay}.} \bibinfo{year}{2014}\natexlab{}.
\newblock \showarticletitle{Large dimensional analysis and optimization of
  robust shrinkage covariance matrix estimators}.
\newblock \bibinfo{journal}{\emph{J. Multivar. Anal.}}  \bibinfo{volume}{131}
  (\bibinfo{year}{2014}), \bibinfo{pages}{99--120}.
\newblock


\bibitem[\protect\citeauthoryear{Daniusis and Vaitkus}{Daniusis and
  Vaitkus}{2008}]%
        {Daniusis2008NeuralNW}
\bibfield{author}{\bibinfo{person}{Povilas Daniusis} {and}
  \bibinfo{person}{Pranas Vaitkus}.} \bibinfo{year}{2008}\natexlab{}.
\newblock \showarticletitle{Neural Network with Matrix Inputs}.
\newblock \bibinfo{journal}{\emph{Informatica}}  \bibinfo{volume}{19}
  (\bibinfo{year}{2008}), \bibinfo{pages}{477--486}.
\newblock


\bibitem[\protect\citeauthoryear{Dauphin, Fan, Auli, and Grangier}{Dauphin
  et~al\mbox{.}}{2017}]%
        {Dauphin2017LanguageMW}
\bibfield{author}{\bibinfo{person}{Yann Dauphin}, \bibinfo{person}{Angela Fan},
  \bibinfo{person}{Michael Auli}, {and} \bibinfo{person}{David Grangier}.}
  \bibinfo{year}{2017}\natexlab{}.
\newblock \showarticletitle{Language Modeling with Gated Convolutional
  Networks}. In \bibinfo{booktitle}{\emph{ICML}}.
\newblock


\bibitem[\protect\citeauthoryear{Donoho, Gavish, and Johnstone}{Donoho
  et~al\mbox{.}}{2018}]%
        {Donoho2018OptimalSO}
\bibfield{author}{\bibinfo{person}{David~L. Donoho}, \bibinfo{person}{Matan
  Gavish}, {and} \bibinfo{person}{Iain~M. Johnstone}.}
  \bibinfo{year}{2018}\natexlab{}.
\newblock \showarticletitle{Optimal Shrinkage of Eigenvalues in the Spiked
  Covariance Model.}
\newblock \bibinfo{journal}{\emph{Annals of statistics}}  \bibinfo{volume}{46
  4} (\bibinfo{year}{2018}), \bibinfo{pages}{1742--1778}.
\newblock


\bibitem[\protect\citeauthoryear{Franceschi, Dieuleveut, and Jaggi}{Franceschi
  et~al\mbox{.}}{2019}]%
        {Franceschi2019UnsupervisedSR}
\bibfield{author}{\bibinfo{person}{Jean-Yves Franceschi},
  \bibinfo{person}{Aymeric Dieuleveut}, {and} \bibinfo{person}{Martin Jaggi}.}
  \bibinfo{year}{2019}\natexlab{}.
\newblock \showarticletitle{Unsupervised Scalable Representation Learning for
  Multivariate Time Series}. In \bibinfo{booktitle}{\emph{NeurIPS}}.
\newblock


\bibitem[\protect\citeauthoryear{Friedman, Hastie, and Tibshirani}{Friedman
  et~al\mbox{.}}{2008}]%
        {CompNet9}
\bibfield{author}{\bibinfo{person}{J. Friedman}, \bibinfo{person}{T. Hastie},
  {and} \bibinfo{person}{R. Tibshirani}.} \bibinfo{year}{2008}\natexlab{}.
\newblock \showarticletitle{Sparse inverse covariance estimation with the
  graphical lasso.}
\newblock \bibinfo{journal}{\emph{Biostatistics}}  \bibinfo{volume}{9 3}
  (\bibinfo{year}{2008}), \bibinfo{pages}{432--41}.
\newblock


\bibitem[\protect\citeauthoryear{Fritz, Dorigatti, and R{\"u}gamer}{Fritz
  et~al\mbox{.}}{2021}]%
        {Fritz2021CombiningGN}
\bibfield{author}{\bibinfo{person}{Cornelius Fritz}, \bibinfo{person}{Emilio
  Dorigatti}, {and} \bibinfo{person}{D. R{\"u}gamer}.}
  \bibinfo{year}{2021}\natexlab{}.
\newblock \showarticletitle{Combining Graph Neural Networks and Spatio-temporal
  Disease Models to Predict COVID-19 Cases in Germany}.
\newblock \bibinfo{journal}{\emph{ArXiv}}  \bibinfo{volume}{abs/2101.00661}
  (\bibinfo{year}{2021}).
\newblock


\bibitem[\protect\citeauthoryear{Gao, Guo, and Wang}{Gao et~al\mbox{.}}{2017}]%
        {Gao2017MatrixNN}
\bibfield{author}{\bibinfo{person}{Junbin Gao}, \bibinfo{person}{Yi Guo}, {and}
  \bibinfo{person}{Zhiyong Wang}.} \bibinfo{year}{2017}\natexlab{}.
\newblock \showarticletitle{Matrix Neural Networks}.
\newblock \bibinfo{journal}{\emph{ArXiv}}  \bibinfo{volume}{abs/1601.03805}
  (\bibinfo{year}{2017}).
\newblock


\bibitem[\protect\citeauthoryear{Gatta, Moscato, Postiglione, and
  Sperl{\'i}}{Gatta et~al\mbox{.}}{2021}]%
        {LaGatta2021AnEN}
\bibfield{author}{\bibinfo{person}{Valerio~La Gatta}, \bibinfo{person}{Vincenzo
  Moscato}, \bibinfo{person}{Marco Postiglione}, {and}
  \bibinfo{person}{Giancarlo Sperl{\'i}}.} \bibinfo{year}{2021}\natexlab{}.
\newblock \showarticletitle{An Epidemiological Neural Network Exploiting
  Dynamic Graph Structured Data Applied to the COVID-19 Outbreak}.
\newblock \bibinfo{journal}{\emph{IEEE Transactions on Big Data}}
  \bibinfo{volume}{7} (\bibinfo{year}{2021}), \bibinfo{pages}{45--55}.
\newblock


\bibitem[\protect\citeauthoryear{Grassia and Mangioni}{Grassia and
  Mangioni}{2022}]%
        {Grassia2022wsGATWA}
\bibfield{author}{\bibinfo{person}{Marco Grassia} {and}
  \bibinfo{person}{Giuseppe Mangioni}.} \bibinfo{year}{2022}\natexlab{}.
\newblock \showarticletitle{wsGAT: Weighted and Signed Graph Attention Networks
  for Link Prediction}.
\newblock \bibinfo{journal}{\emph{ArXiv}}  \bibinfo{volume}{abs/2109.11519}
  (\bibinfo{year}{2022}).
\newblock


\bibitem[\protect\citeauthoryear{Harris, Millman, van~der Walt, Gommers,
  Virtanen, Cournapeau, Wieser, Taylor, Berg, Smith, Kern, Picus, Hoyer, van
  Kerkwijk, Brett, Haldane, Fernández~del Río, Wiebe, Peterson,
  Gérard-Marchant, Sheppard, Reddy, Weckesser, Abbasi, Gohlke, and
  Oliphant}{Harris et~al\mbox{.}}{2020}]%
        {2020NumPy-Array}
\bibfield{author}{\bibinfo{person}{Charles~R. Harris},
  \bibinfo{person}{K.~Jarrod Millman}, \bibinfo{person}{Stéfan~J van~der
  Walt}, \bibinfo{person}{Ralf Gommers}, \bibinfo{person}{Pauli Virtanen},
  \bibinfo{person}{David Cournapeau}, \bibinfo{person}{Eric Wieser},
  \bibinfo{person}{Julian Taylor}, \bibinfo{person}{Sebastian Berg},
  \bibinfo{person}{Nathaniel~J. Smith}, \bibinfo{person}{Robert Kern},
  \bibinfo{person}{Matti Picus}, \bibinfo{person}{Stephan Hoyer},
  \bibinfo{person}{Marten~H. van Kerkwijk}, \bibinfo{person}{Matthew Brett},
  \bibinfo{person}{Allan Haldane}, \bibinfo{person}{Jaime Fernández~del Río},
  \bibinfo{person}{Mark Wiebe}, \bibinfo{person}{Pearu Peterson},
  \bibinfo{person}{Pierre Gérard-Marchant}, \bibinfo{person}{Kevin Sheppard},
  \bibinfo{person}{Tyler Reddy}, \bibinfo{person}{Warren Weckesser},
  \bibinfo{person}{Hameer Abbasi}, \bibinfo{person}{Christoph Gohlke}, {and}
  \bibinfo{person}{Travis~E. Oliphant}.} \bibinfo{year}{2020}\natexlab{}.
\newblock \showarticletitle{Array programming with {NumPy}}.
\newblock \bibinfo{journal}{\emph{Nature}}  \bibinfo{volume}{585}
  (\bibinfo{year}{2020}), \bibinfo{pages}{357–362}.
\newblock
\urldef\tempurl%
\url{https://doi.org/10.1038/s41586-020-2649-2}
\showDOI{\tempurl}


\bibitem[\protect\citeauthoryear{Hasanzadeh, Hajiramezanali, Boluki, Zhou,
  Duffield, Narayanan, and Qian}{Hasanzadeh et~al\mbox{.}}{2020}]%
        {Hasanzadeh2020BayesianGN}
\bibfield{author}{\bibinfo{person}{Arman Hasanzadeh}, \bibinfo{person}{Ehsan
  Hajiramezanali}, \bibinfo{person}{Shahin Boluki}, \bibinfo{person}{Mingyuan
  Zhou}, \bibinfo{person}{Nick~G. Duffield}, \bibinfo{person}{Krishna~R.
  Narayanan}, {and} \bibinfo{person}{Xiaoning Qian}.}
  \bibinfo{year}{2020}\natexlab{}.
\newblock \showarticletitle{Bayesian Graph Neural Networks with Adaptive
  Connection Sampling}.
\newblock \bibinfo{journal}{\emph{ArXiv}}  \bibinfo{volume}{abs/2006.04064}
  (\bibinfo{year}{2020}).
\newblock


\bibitem[\protect\citeauthoryear{Hou, Wang, Zhong, and Wei}{Hou
  et~al\mbox{.}}{2021}]%
        {Hou2021STTraderAS}
\bibfield{author}{\bibinfo{person}{Xiurui Hou}, \bibinfo{person}{Kai Wang},
  \bibinfo{person}{Cheng Zhong}, {and} \bibinfo{person}{Zhi Wei}.}
  \bibinfo{year}{2021}\natexlab{}.
\newblock \showarticletitle{ST-Trader: A Spatial-Temporal Deep Neural Network
  for Modeling Stock Market Movement}.
\newblock \bibinfo{journal}{\emph{IEEE/CAA Journal of Automatica Sinica}}
  \bibinfo{volume}{8} (\bibinfo{year}{2021}), \bibinfo{pages}{1015--1024}.
\newblock


\bibitem[\protect\citeauthoryear{Kaggle}{Kaggle}{[n.d.]}]%
        {kaggle}
\bibfield{author}{\bibinfo{person}{Kaggle}.} \bibinfo{year}{[n.d.]}\natexlab{}.
\newblock \bibinfo{title}{Kaggle: Store Item Demand Forecasting Challenge}.
\newblock
  \bibinfo{howpublished}{\url{https://www.kaggle.com/c/demand-forecasting-kernels-only/data}}.
\newblock
\newblock
\shownote{Accessed: 2022-02-06.}


\bibitem[\protect\citeauthoryear{Kapoor, Ben, Liu, Perozzi, Barnes, Blais, and
  O’Banion}{Kapoor et~al\mbox{.}}{2020}]%
        {Kapoor2020ExaminingCF}
\bibfield{author}{\bibinfo{person}{Amol Kapoor}, \bibinfo{person}{Xue Ben},
  \bibinfo{person}{Luyang Liu}, \bibinfo{person}{Bryan Perozzi},
  \bibinfo{person}{Matt Barnes}, \bibinfo{person}{Martin~J. Blais}, {and}
  \bibinfo{person}{Shawn O’Banion}.} \bibinfo{year}{2020}\natexlab{}.
\newblock \showarticletitle{Examining COVID-19 Forecasting using
  Spatio-Temporal Graph Neural Networks}.
\newblock \bibinfo{journal}{\emph{ArXiv}}  \bibinfo{volume}{abs/2007.03113}
  (\bibinfo{year}{2020}).
\newblock


\bibitem[\protect\citeauthoryear{Kascha}{Kascha}{2012}]%
        {Kascha2012ACO}
\bibfield{author}{\bibinfo{person}{Christian Kascha}.}
  \bibinfo{year}{2012}\natexlab{}.
\newblock \showarticletitle{A Comparison of Estimation Methods for Vector
  Autoregressive Moving-Average Models}.
\newblock \bibinfo{journal}{\emph{Econometric Reviews}}  \bibinfo{volume}{31}
  (\bibinfo{year}{2012}), \bibinfo{pages}{297 -- 324}.
\newblock


\bibitem[\protect\citeauthoryear{Khodayar and Wang}{Khodayar and Wang}{2019}]%
        {Khodayar2019SpatioTemporalGD}
\bibfield{author}{\bibinfo{person}{Mahdi Khodayar} {and}
  \bibinfo{person}{Jianhui Wang}.} \bibinfo{year}{2019}\natexlab{}.
\newblock \showarticletitle{Spatio-Temporal Graph Deep Neural Network for
  Short-Term Wind Speed Forecasting}.
\newblock \bibinfo{journal}{\emph{IEEE Transactions on Sustainable Energy}}
  \bibinfo{volume}{10} (\bibinfo{year}{2019}), \bibinfo{pages}{670--681}.
\newblock


\bibitem[\protect\citeauthoryear{Kim and Oh}{Kim and Oh}{2021}]%
        {Kim2021HowTF}
\bibfield{author}{\bibinfo{person}{Dongkwan Kim} {and}
  \bibinfo{person}{Alice~H. Oh}.} \bibinfo{year}{2021}\natexlab{}.
\newblock \showarticletitle{How to Find Your Friendly Neighborhood: Graph
  Attention Design with Self-Supervision}. In \bibinfo{booktitle}{\emph{ICLR}}.
\newblock


\bibitem[\protect\citeauthoryear{Kipf and Welling}{Kipf and Welling}{2017}]%
        {Kipf2017SemiSupervisedCW}
\bibfield{author}{\bibinfo{person}{Thomas Kipf} {and} \bibinfo{person}{Max
  Welling}.} \bibinfo{year}{2017}\natexlab{}.
\newblock \showarticletitle{Semi-Supervised Classification with Graph
  Convolutional Networks}.
\newblock \bibinfo{journal}{\emph{ArXiv}}  \bibinfo{volume}{abs/1609.02907}
  (\bibinfo{year}{2017}).
\newblock


\bibitem[\protect\citeauthoryear{Kojaku and Masuda}{Kojaku and Masuda}{2019}]%
        {Kojaku2019ConstructingNB}
\bibfield{author}{\bibinfo{person}{Sadamori Kojaku} {and}
  \bibinfo{person}{Naoki Masuda}.} \bibinfo{year}{2019}\natexlab{}.
\newblock \showarticletitle{Constructing networks by filtering correlation
  matrices: a null model approach}.
\newblock \bibinfo{journal}{\emph{Proceedings of the Royal Society A}}
  \bibinfo{volume}{475} (\bibinfo{year}{2019}).
\newblock


\bibitem[\protect\citeauthoryear{Kruskal}{Kruskal}{1956}]%
        {CompNet10}
\bibfield{author}{\bibinfo{person}{Joseph~B. Kruskal}.}
  \bibinfo{year}{1956}\natexlab{}.
\newblock \showarticletitle{On the shortest spanning subtree of a graph and the
  traveling salesman problem}.
\newblock


\bibitem[\protect\citeauthoryear{Lai, Chang, Yang, and Liu}{Lai
  et~al\mbox{.}}{2018}]%
        {Lai2018ModelingLA}
\bibfield{author}{\bibinfo{person}{Guokun Lai}, \bibinfo{person}{Wei-Cheng
  Chang}, \bibinfo{person}{Yiming Yang}, {and} \bibinfo{person}{Hanxiao Liu}.}
  \bibinfo{year}{2018}\natexlab{}.
\newblock \showarticletitle{Modeling Long- and Short-Term Temporal Patterns
  with Deep Neural Networks}.
\newblock \bibinfo{journal}{\emph{The 41st International ACM SIGIR Conference
  on Research \& Development in Information Retrieval}} (\bibinfo{year}{2018}).
\newblock


\bibitem[\protect\citeauthoryear{Ledoit and Wolf}{Ledoit and Wolf}{2003}]%
        {Ledoit2003HoneyIS}
\bibfield{author}{\bibinfo{person}{Olivier Ledoit} {and}
  \bibinfo{person}{Michael Wolf}.} \bibinfo{year}{2003}\natexlab{}.
\newblock \showarticletitle{Honey, I Shrunk the Sample Covariance Matrix}.
\newblock \bibinfo{journal}{\emph{Capital Markets: Asset Pricing \& Valuation}}
  (\bibinfo{year}{2003}).
\newblock


\bibitem[\protect\citeauthoryear{Ledoit and Wolf}{Ledoit and Wolf}{2004}]%
        {Ledoit2004AWE}
\bibfield{author}{\bibinfo{person}{Olivier Ledoit} {and}
  \bibinfo{person}{Michael Wolf}.} \bibinfo{year}{2004}\natexlab{}.
\newblock \showarticletitle{A well-conditioned estimator for large-dimensional
  covariance matrices}.
\newblock \bibinfo{journal}{\emph{Journal of Multivariate Analysis}}
  \bibinfo{volume}{88} (\bibinfo{year}{2004}), \bibinfo{pages}{365--411}.
\newblock


\bibitem[\protect\citeauthoryear{Ledoit and Wolf}{Ledoit and Wolf}{2012}]%
        {Ledoit2012NonlinearSE}
\bibfield{author}{\bibinfo{person}{Olivier Ledoit} {and}
  \bibinfo{person}{Michael Wolf}.} \bibinfo{year}{2012}\natexlab{}.
\newblock \showarticletitle{Nonlinear shrinkage estimation of large-dimensional
  covariance matrices}.
\newblock \bibinfo{journal}{\emph{arXiv: Statistics Theory}}
  (\bibinfo{year}{2012}).
\newblock


\bibitem[\protect\citeauthoryear{Lee and Seregina}{Lee and Seregina}{2020}]%
        {Lee2020OptimalPU}
\bibfield{author}{\bibinfo{person}{Tae-Hwy Lee} {and}
  \bibinfo{person}{Ekaterina Seregina}.} \bibinfo{year}{2020}\natexlab{}.
\newblock \showarticletitle{Optimal Portfolio Using Factor Graphical Lasso}.
\newblock \bibinfo{journal}{\emph{arXiv: Econometrics}} (\bibinfo{year}{2020}).
\newblock


\bibitem[\protect\citeauthoryear{Li, Yu, Shahabi, and Liu}{Li
  et~al\mbox{.}}{2018}]%
        {Li2018DiffusionCR}
\bibfield{author}{\bibinfo{person}{Yaguang Li}, \bibinfo{person}{Rose Yu},
  \bibinfo{person}{Cyrus Shahabi}, {and} \bibinfo{person}{Yan Liu}.}
  \bibinfo{year}{2018}\natexlab{}.
\newblock \showarticletitle{Diffusion Convolutional Recurrent Neural Network:
  Data-Driven Traffic Forecasting}.
\newblock \bibinfo{journal}{\emph{arXiv: Learning}} (\bibinfo{year}{2018}).
\newblock


\bibitem[\protect\citeauthoryear{Luo, Cheng, Yu, Zong, Ni, Chen, and Zhang}{Luo
  et~al\mbox{.}}{2021}]%
        {Luo2021LearningTD}
\bibfield{author}{\bibinfo{person}{Dongsheng Luo}, \bibinfo{person}{Wei Cheng},
  \bibinfo{person}{Wenchao Yu}, \bibinfo{person}{Bo Zong},
  \bibinfo{person}{Jingchao Ni}, \bibinfo{person}{Haifeng Chen}, {and}
  \bibinfo{person}{Xiang Zhang}.} \bibinfo{year}{2021}\natexlab{}.
\newblock \showarticletitle{Learning to Drop: Robust Graph Neural Network via
  Topological Denoising}.
\newblock \bibinfo{journal}{\emph{Proceedings of the 14th ACM International
  Conference on Web Search and Data Mining}} (\bibinfo{year}{2021}).
\newblock


\bibitem[\protect\citeauthoryear{{Markowitz}}{{Markowitz}}{1952}]%
        {Markowitz}
\bibfield{author}{\bibinfo{person}{H. {Markowitz}}.}
  \bibinfo{year}{1952}\natexlab{}.
\newblock \showarticletitle{Portfolio Selection}.
\newblock \bibinfo{journal}{\emph{The Journal of Finance}} \bibinfo{volume}{7},
  \bibinfo{number}{1} (\bibinfo{year}{1952}).
\newblock


\bibitem[\protect\citeauthoryear{Massara and Aste}{Massara and Aste}{2019a}]%
        {CompNet6}
\bibfield{author}{\bibinfo{person}{Guido~Previde Massara} {and}
  \bibinfo{person}{Tomaso Aste}.} \bibinfo{year}{2019}\natexlab{a}.
\newblock \showarticletitle{Learning Clique Forests}.
\newblock \bibinfo{journal}{\emph{ArXiv}}  \bibinfo{volume}{1905.02266}
  (\bibinfo{year}{2019}).
\newblock


\bibitem[\protect\citeauthoryear{Massara and Aste}{Massara and Aste}{2019b}]%
        {Massara2019LearningCF}
\bibfield{author}{\bibinfo{person}{Guido~Previde Massara} {and}
  \bibinfo{person}{Tomaso Aste}.} \bibinfo{year}{2019}\natexlab{b}.
\newblock \showarticletitle{Learning Clique Forests}.
\newblock \bibinfo{journal}{\emph{ArXiv}}  \bibinfo{volume}{abs/1905.02266}
  (\bibinfo{year}{2019}).
\newblock


\bibitem[\protect\citeauthoryear{Massara, Matteo, and Aste}{Massara
  et~al\mbox{.}}{2017}]%
        {CompNet5}
\bibfield{author}{\bibinfo{person}{Guido~Previde Massara}, \bibinfo{person}{T.
  Matteo}, {and} \bibinfo{person}{T. Aste}.} \bibinfo{year}{2017}\natexlab{}.
\newblock \showarticletitle{Network Filtering for Big Data: Triangulated
  Maximally Filtered Graph}.
\newblock \bibinfo{journal}{\emph{ArXiv}}  \bibinfo{volume}{abs/1505.02445}
  (\bibinfo{year}{2017}).
\newblock


\bibitem[\protect\citeauthoryear{Matsunaga, Suzumura, and Takahashi}{Matsunaga
  et~al\mbox{.}}{2019}]%
        {Matsunaga2019ExploringGN}
\bibfield{author}{\bibinfo{person}{Daiki Matsunaga}, \bibinfo{person}{Toyotaro
  Suzumura}, {and} \bibinfo{person}{Toshihiro Takahashi}.}
  \bibinfo{year}{2019}\natexlab{}.
\newblock \showarticletitle{Exploring Graph Neural Networks for Stock Market
  Predictions with Rolling Window Analysis}.
\newblock \bibinfo{journal}{\emph{ArXiv}}  \bibinfo{volume}{abs/1909.10660}
  (\bibinfo{year}{2019}).
\newblock


\bibitem[\protect\citeauthoryear{Mazumder and Hastie}{Mazumder and
  Hastie}{2012}]%
        {Mazumder2012TheGL}
\bibfield{author}{\bibinfo{person}{Rahul Mazumder} {and}
  \bibinfo{person}{Trevor~J. Hastie}.} \bibinfo{year}{2012}\natexlab{}.
\newblock \showarticletitle{The Graphical Lasso: New Insights and
  Alternatives}.
\newblock \bibinfo{journal}{\emph{Electronic journal of statistics}}
  \bibinfo{volume}{6} (\bibinfo{year}{2012}), \bibinfo{pages}{2125--2149}.
\newblock


\bibitem[\protect\citeauthoryear{Meinshausen and Buhlmann}{Meinshausen and
  Buhlmann}{2006}]%
        {Meinshausen2006HighdimensionalGA}
\bibfield{author}{\bibinfo{person}{Nicolai Meinshausen} {and}
  \bibinfo{person}{Peter Buhlmann}.} \bibinfo{year}{2006}\natexlab{}.
\newblock \showarticletitle{High-dimensional graphs and variable selection with
  the Lasso}.
\newblock \bibinfo{journal}{\emph{Annals of Statistics}}  \bibinfo{volume}{34}
  (\bibinfo{year}{2006}), \bibinfo{pages}{1436--1462}.
\newblock


\bibitem[\protect\citeauthoryear{Millington and Niranjan}{Millington and
  Niranjan}{2017}]%
        {Millington2017RobustPR}
\bibfield{author}{\bibinfo{person}{Tristan Millington} {and}
  \bibinfo{person}{Mahesan Niranjan}.} \bibinfo{year}{2017}\natexlab{}.
\newblock \showarticletitle{Robust Portfolio Risk Minimization Using the
  Graphical Lasso}. In \bibinfo{booktitle}{\emph{ICONIP}}.
\newblock


\bibitem[\protect\citeauthoryear{Ne{\v{s}}et{\v{r}}il, Milkov{\'a}, and
  Ne{\v{s}}et{\v{r}}ilov{\'a}}{Ne{\v{s}}et{\v{r}}il et~al\mbox{.}}{2001}]%
        {nevsetvril2001otakar}
\bibfield{author}{\bibinfo{person}{Jaroslav Ne{\v{s}}et{\v{r}}il},
  \bibinfo{person}{Eva Milkov{\'a}}, {and} \bibinfo{person}{Helena
  Ne{\v{s}}et{\v{r}}ilov{\'a}}.} \bibinfo{year}{2001}\natexlab{}.
\newblock \showarticletitle{Otakar Boruvka on minimum spanning tree problem
  Translation of both the 1926 papers, comments, history}.
\newblock \bibinfo{journal}{\emph{Discrete mathematics}} \bibinfo{volume}{233},
  \bibinfo{number}{1-3} (\bibinfo{year}{2001}), \bibinfo{pages}{3--36}.
\newblock


\bibitem[\protect\citeauthoryear{Pacreau, Lezmi, and Xu}{Pacreau
  et~al\mbox{.}}{2021}]%
        {Amudi}
\bibfield{author}{\bibinfo{person}{Grégoire Pacreau}, \bibinfo{person}{Edmond
  Lezmi}, {and} \bibinfo{person}{Jiali Xu}.} \bibinfo{year}{2021}\natexlab{}.
\newblock \showarticletitle{Graph Neural Networks for Asset Management}.
\newblock \bibinfo{journal}{\emph{SSRN}} (\bibinfo{year}{2021}).
\newblock


\bibitem[\protect\citeauthoryear{Pedregosa, Varoquaux, Gramfort, Michel,
  Thirion, Grisel, Blondel, Prettenhofer, Weiss, Dubourg, Vanderplas, Passos,
  Cournapeau, Brucher, Perrot, and Duchesnay}{Pedregosa et~al\mbox{.}}{2011}]%
        {scikit-learn}
\bibfield{author}{\bibinfo{person}{F. Pedregosa}, \bibinfo{person}{G.
  Varoquaux}, \bibinfo{person}{A. Gramfort}, \bibinfo{person}{V. Michel},
  \bibinfo{person}{B. Thirion}, \bibinfo{person}{O. Grisel},
  \bibinfo{person}{M. Blondel}, \bibinfo{person}{P. Prettenhofer},
  \bibinfo{person}{R. Weiss}, \bibinfo{person}{V. Dubourg}, \bibinfo{person}{J.
  Vanderplas}, \bibinfo{person}{A. Passos}, \bibinfo{person}{D. Cournapeau},
  \bibinfo{person}{M. Brucher}, \bibinfo{person}{M. Perrot}, {and}
  \bibinfo{person}{E. Duchesnay}.} \bibinfo{year}{2011}\natexlab{}.
\newblock \showarticletitle{Scikit-learn: Machine Learning in {P}ython}.
\newblock \bibinfo{journal}{\emph{Journal of Machine Learning Research}}
  \bibinfo{volume}{12} (\bibinfo{year}{2011}), \bibinfo{pages}{2825--2830}.
\newblock


\bibitem[\protect\citeauthoryear{Pillay and Moodley}{Pillay and
  Moodley}{2022}]%
        {Pillay2022ExploringGN}
\bibfield{author}{\bibinfo{person}{Kialan Pillay} {and}
  \bibinfo{person}{Deshendran Moodley}.} \bibinfo{year}{2022}\natexlab{}.
\newblock \showarticletitle{Exploring Graph Neural Networks for Stock Market
  Prediction on the JSE}.
\newblock \bibinfo{journal}{\emph{Artificial Intelligence Research}}
  (\bibinfo{year}{2022}).
\newblock


\bibitem[\protect\citeauthoryear{Prim}{Prim}{1957}]%
        {CompNet11}
\bibfield{author}{\bibinfo{person}{Robert~C. Prim}.}
  \bibinfo{year}{1957}\natexlab{}.
\newblock \showarticletitle{Shortest connection networks and some
  generalizations}.
\newblock \bibinfo{journal}{\emph{Bell System Technical Journal}}
  \bibinfo{volume}{36} (\bibinfo{year}{1957}), \bibinfo{pages}{1389--1401}.
\newblock


\bibitem[\protect\citeauthoryear{Procacci and Aste}{Procacci and Aste}{2021}]%
        {Procacci2021ForecastingMS}
\bibfield{author}{\bibinfo{person}{Pier~Francesco Procacci} {and}
  \bibinfo{person}{Tomaso Aste}.} \bibinfo{year}{2021}\natexlab{}.
\newblock \showarticletitle{Forecasting market states}.
\newblock \bibinfo{journal}{\emph{Machine Learning and AI in Finance}}
  (\bibinfo{year}{2021}).
\newblock


\bibitem[\protect\citeauthoryear{Rangapuram, Seeger, Gasthaus, Stella, Wang,
  and Januschowski}{Rangapuram et~al\mbox{.}}{2018}]%
        {Rangapuram2018DeepSS}
\bibfield{author}{\bibinfo{person}{Syama~Sundar Rangapuram},
  \bibinfo{person}{Matthias~W. Seeger}, \bibinfo{person}{Jan Gasthaus},
  \bibinfo{person}{Lorenzo Stella}, \bibinfo{person}{Bernie Wang}, {and}
  \bibinfo{person}{Tim Januschowski}.} \bibinfo{year}{2018}\natexlab{}.
\newblock \showarticletitle{Deep State Space Models for Time Series
  Forecasting}. In \bibinfo{booktitle}{\emph{NeurIPS}}.
\newblock


\bibitem[\protect\citeauthoryear{Rong, bing Huang, Xu, and Huang}{Rong
  et~al\mbox{.}}{2020}]%
        {Rong2020DropEdgeTD}
\bibfield{author}{\bibinfo{person}{Yu Rong}, \bibinfo{person}{Wen bing Huang},
  \bibinfo{person}{Tingyang Xu}, {and} \bibinfo{person}{Junzhou Huang}.}
  \bibinfo{year}{2020}\natexlab{}.
\newblock \showarticletitle{DropEdge: Towards Deep Graph Convolutional Networks
  on Node Classification}. In \bibinfo{booktitle}{\emph{ICLR}}.
\newblock


\bibitem[\protect\citeauthoryear{Rumelhart, Hinton, and Williams}{Rumelhart
  et~al\mbox{.}}{1986}]%
        {Rumelhart1986LearningIR}
\bibfield{author}{\bibinfo{person}{David~E. Rumelhart},
  \bibinfo{person}{Geoffrey~E. Hinton}, {and} \bibinfo{person}{Ronald~J.
  Williams}.} \bibinfo{year}{1986}\natexlab{}.
\newblock \showarticletitle{Learning internal representations by error
  propagation}.
\newblock


\bibitem[\protect\citeauthoryear{Sak, Senior, and Beaufays}{Sak
  et~al\mbox{.}}{2014}]%
        {Sak2014LongSM}
\bibfield{author}{\bibinfo{person}{Hasim Sak}, \bibinfo{person}{Andrew~W.
  Senior}, {and} \bibinfo{person}{Françoise Beaufays}.}
  \bibinfo{year}{2014}\natexlab{}.
\newblock \showarticletitle{Long short-term memory recurrent neural network
  architectures for large scale acoustic modeling}. In
  \bibinfo{booktitle}{\emph{INTERSPEECH}}.
\newblock


\bibitem[\protect\citeauthoryear{Sen, Yu, and Dhillon}{Sen
  et~al\mbox{.}}{2019}]%
        {Sen2019ThinkGA}
\bibfield{author}{\bibinfo{person}{Rajat Sen}, \bibinfo{person}{Hsiang-Fu Yu},
  {and} \bibinfo{person}{Inderjit~S. Dhillon}.}
  \bibinfo{year}{2019}\natexlab{}.
\newblock \showarticletitle{Think Globally, Act Locally: A Deep Neural Network
  Approach to High-Dimensional Time Series Forecasting}. In
  \bibinfo{booktitle}{\emph{NeurIPS}}.
\newblock


\bibitem[\protect\citeauthoryear{Sesti, Luis, Crawley, and Cameron}{Sesti
  et~al\mbox{.}}{2021}]%
        {Sesti2021IntegratingLA}
\bibfield{author}{\bibinfo{person}{Nathan Sesti}, \bibinfo{person}{Juan
  Jose~Garau Luis}, \bibinfo{person}{Edward~F. Crawley}, {and}
  \bibinfo{person}{Bruce~G. Cameron}.} \bibinfo{year}{2021}\natexlab{}.
\newblock \showarticletitle{Integrating LSTMs and GNNs for COVID-19
  Forecasting}.
\newblock \bibinfo{journal}{\emph{ArXiv}}  \bibinfo{volume}{abs/2108.10052}
  (\bibinfo{year}{2021}).
\newblock


\bibitem[\protect\citeauthoryear{Shen, qi~Cheng, and xing Fang}{Shen
  et~al\mbox{.}}{2010}]%
        {Shen2010CovarianceCM}
\bibfield{author}{\bibinfo{person}{Huawei Shen}, \bibinfo{person}{Xue qi
  Cheng}, {and} \bibinfo{person}{Bin xing Fang}.}
  \bibinfo{year}{2010}\natexlab{}.
\newblock \showarticletitle{Covariance, correlation matrix, and the multiscale
  community structure of networks.}
\newblock \bibinfo{journal}{\emph{Physical review. E, Statistical, nonlinear,
  and soft matter physics}}  \bibinfo{volume}{82 1 Pt 2}
  (\bibinfo{year}{2010}), \bibinfo{pages}{016114}.
\newblock


\bibitem[\protect\citeauthoryear{Shi, Zhang, Cheng, and Lu}{Shi
  et~al\mbox{.}}{2019}]%
        {Shi2019TwoStreamAG}
\bibfield{author}{\bibinfo{person}{Lei Shi}, \bibinfo{person}{Yifan Zhang},
  \bibinfo{person}{Jian Cheng}, {and} \bibinfo{person}{Hanqing Lu}.}
  \bibinfo{year}{2019}\natexlab{}.
\newblock \showarticletitle{Two-Stream Adaptive Graph Convolutional Networks
  for Skeleton-Based Action Recognition}.
\newblock \bibinfo{journal}{\emph{2019 IEEE/CVF Conference on Computer Vision
  and Pattern Recognition (CVPR)}} (\bibinfo{year}{2019}),
  \bibinfo{pages}{12018--12027}.
\newblock


\bibitem[\protect\citeauthoryear{Shi, Chen, Wang, Yeung, Wong, and chun
  Woo}{Shi et~al\mbox{.}}{2015}]%
        {Shi2015ConvolutionalLN}
\bibfield{author}{\bibinfo{person}{Xingjian Shi}, \bibinfo{person}{Zhourong
  Chen}, \bibinfo{person}{Hao Wang}, \bibinfo{person}{Dit-Yan Yeung},
  \bibinfo{person}{Wai-Kin Wong}, {and} \bibinfo{person}{Wang chun Woo}.}
  \bibinfo{year}{2015}\natexlab{}.
\newblock \showarticletitle{Convolutional LSTM Network: A Machine Learning
  Approach for Precipitation Nowcasting}. In \bibinfo{booktitle}{\emph{NIPS}}.
\newblock


\bibitem[\protect\citeauthoryear{Shih, Sun, and yi~Lee}{Shih
  et~al\mbox{.}}{2019}]%
        {Shih2019TemporalPA}
\bibfield{author}{\bibinfo{person}{Shun-Yao Shih}, \bibinfo{person}{Fan-Keng
  Sun}, {and} \bibinfo{person}{Hung yi Lee}.} \bibinfo{year}{2019}\natexlab{}.
\newblock \showarticletitle{Temporal pattern attention for multivariate time
  series forecasting}.
\newblock \bibinfo{journal}{\emph{Machine Learning}} (\bibinfo{year}{2019}),
  \bibinfo{pages}{1--21}.
\newblock


\bibitem[\protect\citeauthoryear{Telesford, Simpson, Burdette, Hayasaka, and
  Laurienti}{Telesford et~al\mbox{.}}{2011}]%
        {CompNet8}
\bibfield{author}{\bibinfo{person}{Qawi~K. Telesford}, \bibinfo{person}{S.
  Simpson}, \bibinfo{person}{J. Burdette}, \bibinfo{person}{S. Hayasaka}, {and}
  \bibinfo{person}{P. Laurienti}.} \bibinfo{year}{2011}\natexlab{}.
\newblock \showarticletitle{The Brain as a Complex System: Using Network
  Science as a Tool for Understanding the Brain}.
\newblock \bibinfo{journal}{\emph{Brain connectivity}}  \bibinfo{volume}{1 (4)}
  (\bibinfo{year}{2011}), \bibinfo{pages}{295--308}.
\newblock


\bibitem[\protect\citeauthoryear{Tumminello, Aste, Di~Matteo, and
  Mantegna}{Tumminello et~al\mbox{.}}{2005}]%
        {CompNet3}
\bibfield{author}{\bibinfo{person}{M. Tumminello}, \bibinfo{person}{T. Aste},
  \bibinfo{person}{T. Di~Matteo}, {and} \bibinfo{person}{R.~N. Mantegna}.}
  \bibinfo{year}{2005}\natexlab{}.
\newblock \showarticletitle{A tool for filtering information in complex
  systems}.
\newblock \bibinfo{journal}{\emph{Proceedings of the National Academy of
  Sciences}} \bibinfo{volume}{102}, \bibinfo{number}{30}
  (\bibinfo{year}{2005}), \bibinfo{pages}{10421--10426}.
\newblock
\showISSN{0027-8424}
\urldef\tempurl%
\url{https://doi.org/10.1073/pnas.0500298102}
\showDOI{\tempurl}
\showeprint{https://www.pnas.org/content/102/30/10421.full.pdf}


\bibitem[\protect\citeauthoryear{Turiel, Barucca, and Aste}{Turiel
  et~al\mbox{.}}{2020}]%
        {Turiel2020SimplicialPO}
\bibfield{author}{\bibinfo{person}{Jeremy~D. Turiel}, \bibinfo{person}{Paolo
  Barucca}, {and} \bibinfo{person}{Tomaso Aste}.}
  \bibinfo{year}{2020}\natexlab{}.
\newblock \showarticletitle{Simplicial persistence of financial markets:
  filtering, generative processes and portfolio risk}.
\newblock \bibinfo{journal}{\emph{arXiv: Statistical Finance}}
  (\bibinfo{year}{2020}).
\newblock


\bibitem[\protect\citeauthoryear{Vaswani, Shazeer, Parmar, Uszkoreit, Jones,
  Gomez, Kaiser, and Polosukhin}{Vaswani et~al\mbox{.}}{2017}]%
        {Vaswani2017AttentionIA}
\bibfield{author}{\bibinfo{person}{Ashish Vaswani}, \bibinfo{person}{Noam~M.
  Shazeer}, \bibinfo{person}{Niki Parmar}, \bibinfo{person}{Jakob Uszkoreit},
  \bibinfo{person}{Llion Jones}, \bibinfo{person}{Aidan~N. Gomez},
  \bibinfo{person}{Lukasz Kaiser}, {and} \bibinfo{person}{Illia Polosukhin}.}
  \bibinfo{year}{2017}\natexlab{}.
\newblock \showarticletitle{Attention is All you Need}.
\newblock \bibinfo{journal}{\emph{ArXiv}}  \bibinfo{volume}{abs/1706.03762}
  (\bibinfo{year}{2017}).
\newblock


\bibitem[\protect\citeauthoryear{Wan, Mei, Wang, Liu, and Yang}{Wan
  et~al\mbox{.}}{2019}]%
        {Wan2019MultivariateTC}
\bibfield{author}{\bibinfo{person}{Renzhuo Wan}, \bibinfo{person}{Shuping Mei},
  \bibinfo{person}{Jun Wang}, \bibinfo{person}{Min Liu}, {and}
  \bibinfo{person}{F. Yang}.} \bibinfo{year}{2019}\natexlab{}.
\newblock \showarticletitle{Multivariate Temporal Convolutional Network: A Deep
  Neural Networks Approach for Multivariate Time Series Forecasting}.
\newblock \bibinfo{journal}{\emph{Electronics}} (\bibinfo{year}{2019}).
\newblock


\bibitem[\protect\citeauthoryear{Wang and Aste}{Wang and Aste}{2021}]%
        {Wang2021DynamicPO}
\bibfield{author}{\bibinfo{person}{Yuanrong Wang} {and} \bibinfo{person}{Tomaso
  Aste}.} \bibinfo{year}{2021}\natexlab{}.
\newblock \showarticletitle{Dynamic Portfolio Optimization with Inverse
  Covariance Clustering}.
\newblock


\bibitem[\protect\citeauthoryear{Wu, Pan, Long, Jiang, Chang, and Zhang}{Wu
  et~al\mbox{.}}{2020}]%
        {Wu2020ConnectingTD}
\bibfield{author}{\bibinfo{person}{Zonghan Wu}, \bibinfo{person}{Shirui Pan},
  \bibinfo{person}{Guodong Long}, \bibinfo{person}{Jing Jiang},
  \bibinfo{person}{Xiaojun Chang}, {and} \bibinfo{person}{Chengqi Zhang}.}
  \bibinfo{year}{2020}\natexlab{}.
\newblock \showarticletitle{Connecting the Dots: Multivariate Time Series
  Forecasting with Graph Neural Networks}.
\newblock \bibinfo{journal}{\emph{Proceedings of the 26th ACM SIGKDD
  International Conference on Knowledge Discovery \& Data Mining}}
  (\bibinfo{year}{2020}).
\newblock


\bibitem[\protect\citeauthoryear{Wu, Pan, Long, Jiang, and Zhang}{Wu
  et~al\mbox{.}}{2019}]%
        {Wu2019GraphWF}
\bibfield{author}{\bibinfo{person}{Zonghan Wu}, \bibinfo{person}{Shirui Pan},
  \bibinfo{person}{Guodong Long}, \bibinfo{person}{Jing Jiang}, {and}
  \bibinfo{person}{Chengqi Zhang}.} \bibinfo{year}{2019}\natexlab{}.
\newblock \showarticletitle{Graph WaveNet for Deep Spatial-Temporal Graph
  Modeling}. In \bibinfo{booktitle}{\emph{IJCAI}}.
\newblock


\bibitem[\protect\citeauthoryear{Yan, Xiong, and Lin}{Yan
  et~al\mbox{.}}{2018}]%
        {Yan2018SpatialTG}
\bibfield{author}{\bibinfo{person}{Sijie Yan}, \bibinfo{person}{Yuanjun Xiong},
  {and} \bibinfo{person}{Dahua Lin}.} \bibinfo{year}{2018}\natexlab{}.
\newblock \showarticletitle{Spatial Temporal Graph Convolutional Networks for
  Skeleton-Based Action Recognition}.
\newblock \bibinfo{journal}{\emph{ArXiv}}  \bibinfo{volume}{abs/1801.07455}
  (\bibinfo{year}{2018}).
\newblock


\bibitem[\protect\citeauthoryear{Ying, You, Morris, Ren, Hamilton, and
  Leskovec}{Ying et~al\mbox{.}}{2018}]%
        {Ying2018HierarchicalGR}
\bibfield{author}{\bibinfo{person}{Rex Ying}, \bibinfo{person}{Jiaxuan You},
  \bibinfo{person}{Christopher Morris}, \bibinfo{person}{Xiang Ren},
  \bibinfo{person}{William~L. Hamilton}, {and} \bibinfo{person}{Jure
  Leskovec}.} \bibinfo{year}{2018}\natexlab{}.
\newblock \showarticletitle{Hierarchical Graph Representation Learning with
  Differentiable Pooling}.
\newblock \bibinfo{journal}{\emph{ArXiv}}  \bibinfo{volume}{abs/1806.08804}
  (\bibinfo{year}{2018}).
\newblock


\bibitem[\protect\citeauthoryear{Young and Shellswell}{Young and
  Shellswell}{1972}]%
        {Young1972TimeSA}
\bibfield{author}{\bibinfo{person}{P.~J. Young} {and} \bibinfo{person}{Stephen
  Shellswell}.} \bibinfo{year}{1972}\natexlab{}.
\newblock \showarticletitle{Time series analysis, forecasting and control}.
\newblock \bibinfo{journal}{\emph{IEEE Trans. Automat. Control}}
  \bibinfo{volume}{17} (\bibinfo{year}{1972}), \bibinfo{pages}{281--283}.
\newblock


\bibitem[\protect\citeauthoryear{Yu, Yin, and Zhu}{Yu et~al\mbox{.}}{2018}]%
        {Yu2018SpatioTemporalGC}
\bibfield{author}{\bibinfo{person}{Ting Yu}, \bibinfo{person}{Haoteng Yin},
  {and} \bibinfo{person}{Zhanxing Zhu}.} \bibinfo{year}{2018}\natexlab{}.
\newblock \showarticletitle{Spatio-Temporal Graph Convolutional Networks: A
  Deep Learning Framework for Traffic Forecasting}. In
  \bibinfo{booktitle}{\emph{IJCAI}}.
\newblock


\bibitem[\protect\citeauthoryear{Yuan and Lin}{Yuan and Lin}{2007}]%
        {Yuan2007ModelSA}
\bibfield{author}{\bibinfo{person}{Ming Yuan} {and} \bibinfo{person}{Yi Lin}.}
  \bibinfo{year}{2007}\natexlab{}.
\newblock \showarticletitle{Model selection and estimation in the Gaussian
  graphical model}.
\newblock \bibinfo{journal}{\emph{Biometrika}}  \bibinfo{volume}{94}
  (\bibinfo{year}{2007}), \bibinfo{pages}{19--35}.
\newblock


\bibitem[\protect\citeauthoryear{Yuan, Yu, Yin, and Wang}{Yuan
  et~al\mbox{.}}{2020}]%
        {Yuan2020ImprovedLD}
\bibfield{author}{\bibinfo{person}{Xin Yuan}, \bibinfo{person}{Weiqin Yu},
  \bibinfo{person}{Zhixian Yin}, {and} \bibinfo{person}{Guoqiang Wang}.}
  \bibinfo{year}{2020}\natexlab{}.
\newblock \showarticletitle{Improved Large Dynamic Covariance Matrix Estimation
  With Graphical Lasso and Its Application in Portfolio Selection}.
\newblock \bibinfo{journal}{\emph{IEEE Access}}  \bibinfo{volume}{8}
  (\bibinfo{year}{2020}), \bibinfo{pages}{189179--189188}.
\newblock


\bibitem[\protect\citeauthoryear{Zhang, Aggarwal, and Qi}{Zhang
  et~al\mbox{.}}{2017}]%
        {Zhang2017StockPP}
\bibfield{author}{\bibinfo{person}{Liheng Zhang}, \bibinfo{person}{Charu~C.
  Aggarwal}, {and} \bibinfo{person}{Guo-Jun Qi}.}
  \bibinfo{year}{2017}\natexlab{}.
\newblock \showarticletitle{Stock Price Prediction via Discovering
  Multi-Frequency Trading Patterns}.
\newblock \bibinfo{journal}{\emph{Proceedings of the 23rd ACM SIGKDD
  International Conference on Knowledge Discovery and Data Mining}}
  (\bibinfo{year}{2017}).
\newblock


\bibitem[\protect\citeauthoryear{Zhao, Song, Zhang, Liu, Wang, Lin, Deng, and
  Li}{Zhao et~al\mbox{.}}{2020}]%
        {Zhao2020TGCNAT}
\bibfield{author}{\bibinfo{person}{Ling Zhao}, \bibinfo{person}{Yujiao Song},
  \bibinfo{person}{Chao Zhang}, \bibinfo{person}{Yu Liu}, \bibinfo{person}{Pu
  Wang}, \bibinfo{person}{Tao Lin}, \bibinfo{person}{Min Deng}, {and}
  \bibinfo{person}{Haifeng Li}.} \bibinfo{year}{2020}\natexlab{}.
\newblock \showarticletitle{T-GCN: A Temporal Graph Convolutional Network for
  Traffic Prediction}.
\newblock \bibinfo{journal}{\emph{IEEE Transactions on Intelligent
  Transportation Systems}}  \bibinfo{volume}{21} (\bibinfo{year}{2020}),
  \bibinfo{pages}{3848--3858}.
\newblock


\bibitem[\protect\citeauthoryear{Zheng, Fan, Wang, and Qi}{Zheng
  et~al\mbox{.}}{2020}]%
        {Zheng2020GMANAG}
\bibfield{author}{\bibinfo{person}{Chuanpan Zheng}, \bibinfo{person}{Xiaoliang
  Fan}, \bibinfo{person}{Cheng Wang}, {and} \bibinfo{person}{Jianzhong Qi}.}
  \bibinfo{year}{2020}\natexlab{}.
\newblock \showarticletitle{GMAN: A Graph Multi-Attention Network for Traffic
  Prediction}.
\newblock \bibinfo{journal}{\emph{ArXiv}}  \bibinfo{volume}{abs/1911.08415}
  (\bibinfo{year}{2020}).
\newblock


\end{thebibliography}


\end{document}